\documentclass[10pt,twocolumn,letterpaper]{article}

\usepackage[pagenumbers]{cvpr} 

\usepackage{multirow}

\definecolor{cvprblue}{rgb}{0.21,0.49,0.74}
\usepackage[pagebackref,breaklinks,colorlinks,allcolors=cvprblue]{hyperref}
\usepackage{pifont}
\def\paperID{} 
\def\confName{CVPR}
\def\confYear{2026}

\title{Generalized Event Partonomy Inference with Structured Hierarchical Predictive Learning}

\author{Zhou Chen, Joe Lin, Sathyanarayanan N. Aakur\\
CSSE Department, Auburn University\\
Auburn, AL, 36849\\
{\tt\small \{zzc0053,jzl0277,san0028\}@auburn.edu}
}

\begin{document}
\maketitle
\begin{abstract}
Humans naturally perceive continuous experience as a hierarchy of temporally nested events, fine-grained actions embedded within coarser routines. Replicating this structure in computer vision requires models that can segment video not just retrospectively, but predictively and hierarchically. We introduce PARSE, a unified framework that learns multiscale event structure directly from streaming video without supervision. PARSE organizes perception into a hierarchy of recurrent predictors, each operating at its own temporal granularity: lower layers model short-term dynamics while higher layers integrate longer-term context through attention-based feedback. Event boundaries emerge naturally as transient peaks in prediction error, yielding temporally coherent, nested partonomies that mirror the containment relations observed in human event perception. Evaluated across three benchmarks, Breakfast Actions, 50 Salads, and Assembly 101, PARSE achieves state-of-the-art performance among streaming methods and rivals offline baselines in both temporal alignment (H-GEBD) and structural consistency (TED, hF1). The results demonstrate that predictive learning under uncertainty provides a scalable path toward human-like temporal abstraction and compositional event understanding.
Code and evaluation tools will be released publicly upon acceptance.
\end{abstract}

\section{Introduction}\label{sec:intro}
\begin{figure}[t]
    \centering
    \includegraphics[width=0.99\linewidth]{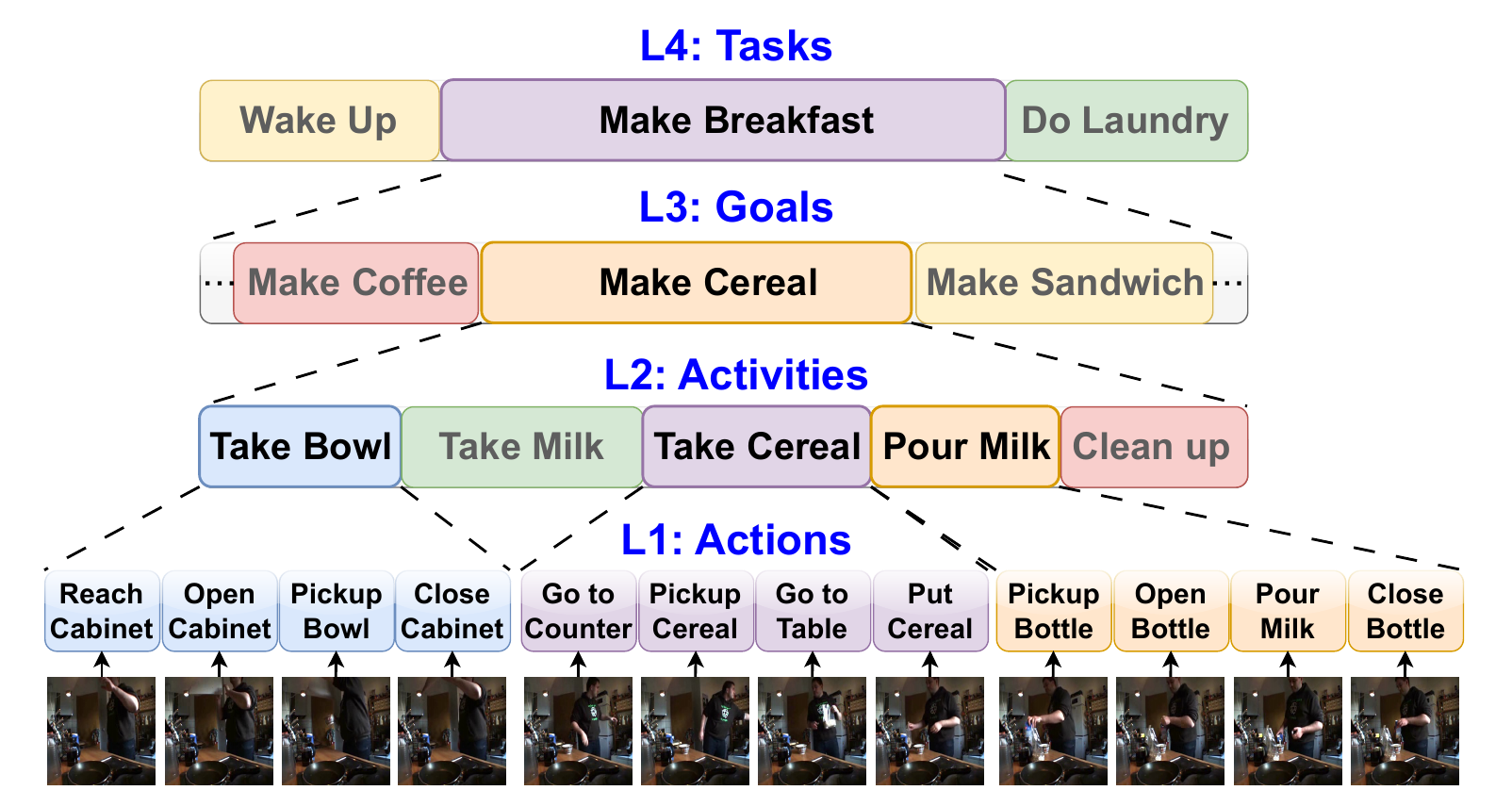}
    \caption{\textbf{Hierarchical Event Partonomy.} Real-world activities unfold across multiple temporal and semantic scales, from long-horizon tasks (L4) decomposed into goals (L3), activities (L2), and atomic actions (L1). Each higher-level subsumes several temporally contained subevents, forming a structured partonomy. 
    }
    \label{fig:intuition}
\end{figure}

Humans experience the world not as a sequence of frames, but as a \textit{structured stream of events} unfolding across \textit{multiple timescales}. When we observe a person making breakfast, we perceive fine-grained actions such as ``\textit{crack egg}'' and ``\textit{pour milk}'' as parts of broader routines, like ``\textit{make omelet}'' or ``\textit{prepare breakfast}''. This nested organization, known as a \textit{partonomy}, is fundamental to how perception, memory, and action interact over time. An example is shown in Figure~\ref{fig:intuition}. Cognitive and neuroscientific theories~\cite{zacks2001event,zacks2001perceiving,tversky2008structure,tversky2014partonomies} suggest that such structure emerges from predictive processing: the brain continually anticipates what happens next, updating internal hypotheses when prediction errors spike. This dynamic coupling of anticipation and reorganization allows humans to segment continuous experience into hierarchically contained events.

Despite significant progress in temporal segmentation and activity recognition~\cite{kuehne2014language,sener2022assembly101,shou2021generic,wang2022geb2,aakur2019perceptual,yi2021asformer,fayyaz2020sct,bansal2024united}, most computer vision models treat event understanding as a flat or retrospective segmentation and labeling problem. Supervised methods~\cite{wang2021end,huang2020improving,sarfraz2021temporally} rely on pre-defined action boundaries and labels, while unsupervised~\cite{aakur2019perceptual,aakur2020action,mounir2023streamer,mounir2023towards} or clustering-based approaches~\cite{sarfraz2021temporally,Kumar_2022_CVPR} segment videos post hoc using global heuristics. These formulations lack the streaming, predictive, and hierarchical characteristics of human perception. They treat segmentation as an after-the-fact optimization rather than an emergent property of how agents anticipate and interpret unfolding experience, resulting in inconsistent boundaries across temporal scales and no notion of \textit{hierarchical containment}.

\begin{figure*}[t]
    \centering
    \includegraphics[width=0.9\linewidth]{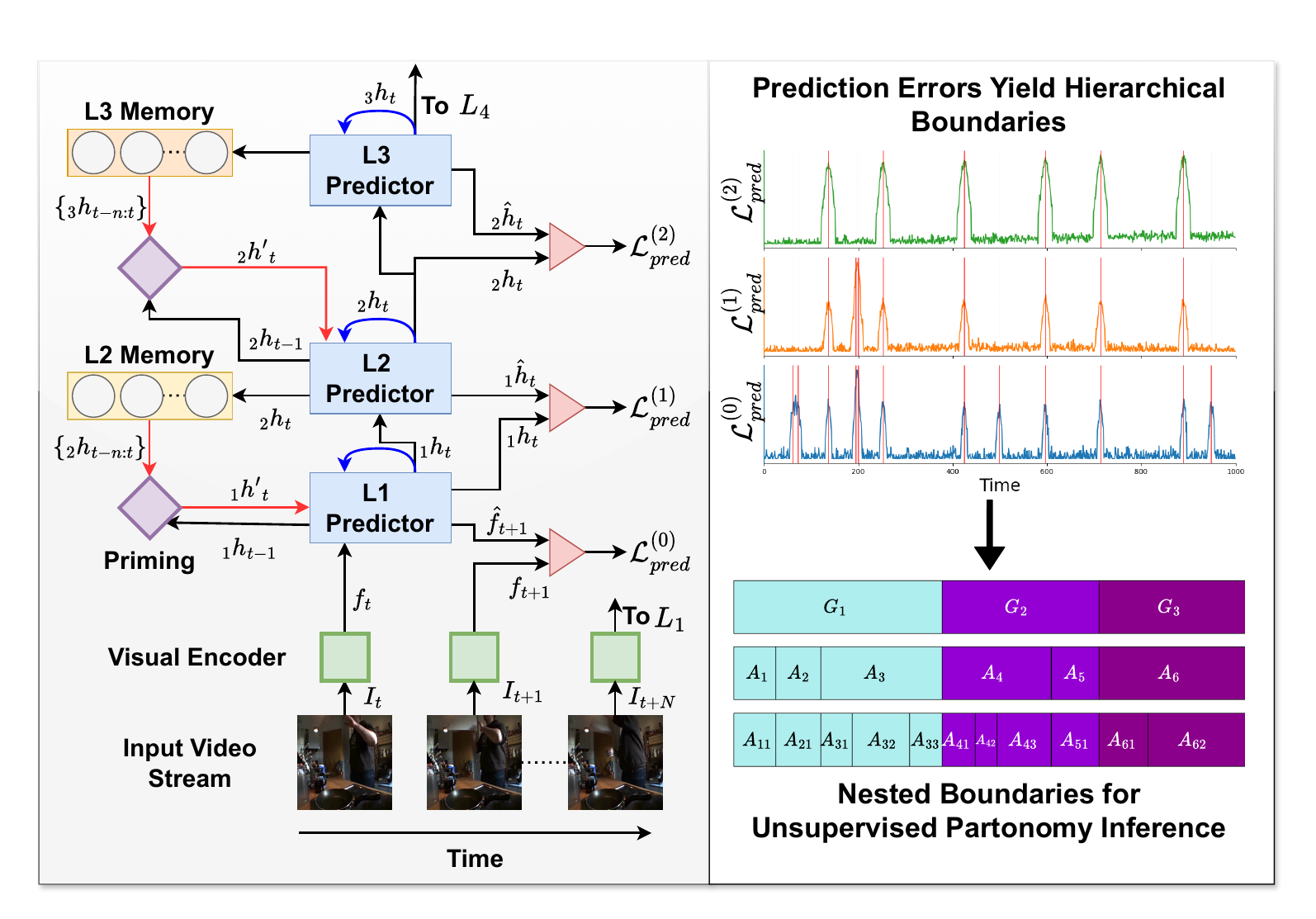}
    \caption{\textbf{PARSE Architecture Overview.}
At each timestep, the visual encoder extracts frame-level features ($f_t$) that feed into a hierarchy of recurrent predictors ($L_1$, $L_2$, $L_3 \ldots L_N$), each modeling temporal dynamics at a distinct scale.
Each predictor generates a forward estimate ($\hat{h}_t^{(i)}$) of its lower-level hidden state ($h_t^{(i-1)}$) and computes a prediction error loss ($\mathcal{L}_{\text{pred}}^{(i)}$).
Transient peaks in these hierarchical prediction errors (right) signal event transitions at progressively coarser temporal scales.
Aggregating these peaks yields \textit{nested event boundaries}, which form a partonomic structure capturing the coarse-to-fine organization of real-world activities.
}
    \label{fig:arch}
\end{figure*}

We propose \textit{PARSE}, a hierarchical predictive learning framework that infers multiscale event structure directly from streaming video. Figure~\ref{fig:arch} illustrates the overall process. Instead of detecting boundaries explicitly, PARSE learns to anticipate future perceptual states at multiple temporal resolutions and reorganizes its internal hypotheses when prediction errors accumulate. Each layer functions as a recurrent predictor operating at its own temporal granularity, with higher layers integrating longer-term dynamics and modulating lower layers through top-down attention. This interplay between bottom-up prediction and top-down contextualization yields temporally aligned, nested boundaries that naturally form a partonomic hierarchy. 

Our \textbf{contributions} are four-fold:
(1) We introduce a formal definition and computational model of hierarchical partonomy inference, framing event segmentation as a process of predictive reorganization.
(2) We develop PARSE, a multi-layer predictive hierarchy that unifies anticipation, sparsity-driven stability, and hierarchical boundary extraction under a single energy-based objective.
(3) We propose an unsupervised boundary inference mechanism that identifies multi-scale transitions from prediction-error dynamics during streaming inference.
(4) We establish evaluation benchmarks for hierarchical event understanding, combining boundary-level (H-GEBD) and structural (TED, hF1) metrics across three datasets, Breakfast Actions~\cite{kuehne2014language}, 50 Salads~\cite{stein2013combining}, and Assembly101~\cite{sener2022assembly101}.


\section{Related Work}\label{sec:prior_work}

\textbf{Predictive learning} conceptualizes perception as an anticipatory process in which internal models continually generate expectations about future sensory input and update their hypotheses when predictions fail. Rooted in cognitive science and neuroscience~\cite{friston2010free,lotter2017deep,clark2013whatever,rao2023active,lupyan2015words}, it posits that perception and attention emerge from minimizing prediction error across hierarchical cortical pathways. Empirical studies of event perception~\cite{zacks2001event,tversky2014partonomies} show that humans segment continuous experience at moments of heightened prediction error, suggesting that event boundaries mark points where generative models of the world are revised. 
Hierarchical predictive inference~\cite{rao2023active,salvatori2023brain,jiang2023dynamic,rao2022sensory} suggests that stable, long-timescale representations can emerge from nested predictive processes. 
In computer vision, this idea has been explored through self-supervised representation learning. Early work~\cite{aakur2019perceptual,wang2023coseg} modeled event segmentation through prediction-error dynamics, while subsequent efforts provide extensions to active control~\cite{trehan2022towards,chen2025ease} and online localization~\cite{aakur2020action,aakur2022actor,trehan2024self,mounir2023streamer,mounir2023towards}. Joint-embedding predictive architectures (JEPA)~\cite{assran2023self,bardes2023mc,littwin2024jepa,bardes2023v,abdelfattah2024s} align temporally adjacent representations in latent space. 

\textbf{Video event understanding} has primarily treated temporal structure as a flat labeling problem, assigning each frame or segment to a single action category. Early approaches used structured temporal models such as HMMs and CRFs~\cite{kuehne2018hybrid,chandra2018deep}, while modern methods employ temporal convolutional networks~\cite{farha2019ms,lea2017temporal}, transformers~\cite{yi2021asformer,fayyaz2020sct}, or graph-based architectures~\cite{huang2020improving,wang2021end} to predict frame-wise actions or segment boundaries. Large-scale benchmarks such as Breakfast Actions~\cite{kuehne2014language}, 50 Salads~\cite{stein2013combining}, EpicKitchens~\cite{damen2018scaling,damen2020epic}, Ego4D~\cite{grauman2022ego4d}, and Assembly101~\cite{sener2022assembly101} have driven rapid progress, along with task variants including action localization~\cite{aakur2020action,aakur2022actor,trehan2024self,mounir2023towards}, procedure segmentation~\cite{ji2022learning,bansal2022my,bansal2024united,zhou2023procedure}, and Generic Event Boundary Detection (GEBD)~\cite{shou2021generic,jung2025online,wang2022geb2}. Yet, these methods largely assume a flat temporal hierarchy, where each frame belongs to one independent class, ignoring the compositional structure that organizes human activity. Even ``hierarchical'' variants that stack models across scales~\cite{sarfraz2021temporally,mounir2023streamer} rely on discrete label taxonomies rather than continuous predictive dynamics. Moreover, nearly all are offline, requiring full-video access before segmentation, limiting their adaptability to streaming or interactive scenarios.

\textbf{Partonomic structure} conceptualizes video understanding as uncovering the nested subevents and containment relations through which activities unfold. Spatial partonomies have been deeply explored in vision, but temporal partonomies remain understudied. Object-centric methods leverage spatial part-whole decompositions for robust recognition and reasoning~\cite{cheng2018fast,zhu2017couplenet,shi2020points,crivellaro2017robust,heisele2001categorization,schneiderman2004object,huber2004parts,leibe2008learning}, while semantic segmentation~\cite{arbelaez2012semantic,zhang2018context,wang2018understanding} and scene-graph frameworks~\cite{krishna2017visual,johnson2015image,ji2020action} capture hierarchical relations among entities within a scene. Extending such insights to the \emph{temporal domain}, where events evolve, overlap, and causally interact, remains an open challenge. Temporal partonomy requires more than boundary detection; it entails identifying \emph{structural containment}, understanding how short, motoric actions compose into longer activities and goals. \textit{PARSE} bridges this gap by learning part-whole hierarchies directly from streaming prediction-error dynamics, producing temporally aligned, multi-scale representations that reflect the latent organization of real-world activities.

\section{Partonomies: Hierarchical Event Structures}
\label{sec:background} 
A \textit{partonomy} captures how complex activities unfold across multiple temporal and semantic scales. Unlike taxonomies, which express \textit{is-a} relationships among symbolic categories, partonomies encode \textit{has-part} relationships among temporally extended subevents that collectively constitute a coherent whole~\cite{tversky2014partonomies}. This representation provides a bridge between low-level perceptual dynamics and high-level task structure, offering a foundation for modeling human-like understanding of long-horizon behavior~\cite{zacks2001event,zacks2001perceiving,tversky2008structure}.

Formally, we define a partonomy as a hierarchy of temporally ordered event segments $\mathcal{P} = \{L_1, L_2, \ldots, L_N\}$, where each level \(L_i = \{e_1^{(i)}, e_2^{(i)}, \ldots, e_{M_i}^{(i)}\}\) consists of disjoint intervals \(e_j^{(i)} = [s_j^{(i)}, t_j^{(i)}]\), where $s_j^{(i)}$ and $t_j^{(i)}$ denote its start and end frame indices, respectively.
Each higher-level event aggregates contiguous lower-level events, enforcing \textit{temporal containment}, as defined by 
\begin{equation}
\mathcal{C}\!\left(e_k^{(i+1)}\right)
= \{\, e_j^{(i)} \in L_i 
\mid s_k^{(i+1)} \le s_j^{(i)} < t_j^{(i)} \le t_k^{(i+1)} \,\}.
\label{eq:containment}
\end{equation}
where $\forall e_k^{(i+1)} \in L_{i+1}$, the operator $\mathcal{C}(\cdot)$ returns its temporally contained children from level $L_i$. 
This containment induces a partial order \(e_a^{(i)} \prec e_b^{(i+1)}\) denoting that \(e_a^{(i)}\) is a constituent of \(e_b^{(i+1)}\), forming a tree-structured decomposition of time.
A valid partonomy therefore satisfies two key properties: (i) \textit{structural coherence}, ensuring consistent parent--child relationships across levels, and (ii) \textit{temporal compositionality}, ensuring that event boundaries are hierarchically aligned rather than independent across scales.

This multiscale organization parallels hierarchical theories of cortical perception, where higher layers integrate evidence over longer timescales while providing contextual feedback to lower ones.
From a computational standpoint, such a hierarchy supports \textit{predictive abstraction}: each level refines its internal hypothesis about ongoing activity through both feedforward evidence and feedback correction, producing a temporally structured representation that is robust to uncertainty, occlusion, and variation. 
Figure~\ref{fig:intuition} illustrates an example partonomy for an everyday routine task such as \textit{making breakfast}. 
Fine-grained motor \textit{actions} (\textit{reach cabinet}, \textit{pour milk}) correspond to level \(L_1\),
mid-level \textit{activities} (\textit{take bowl}, \textit{make cereal}) to level \(L_2\),
and \textit{goal-oriented subroutines} (\textit{prepare breakfast}) to level \(L_3\),
all nested within higher-level tasks (\textit{morning routine}, \textit{daily schedule}) at level \(L_4\).
Each higher layer subsumes multiple temporally contained subevents, capturing both the hierarchical structure and dynamics of human activity. 
In this work, we address the problem of \textit{extracting such hierarchical partonomies directly from streaming video}, without supervision or predefined symbolic labels.
Unlike prior approaches~\cite{mounir2023streamer,mounir2023towards,ji2020action}, we consider the \textit{online, predictive} setting in which an agent must infer the evolving structure of events incrementally as new visual evidence arrives.
This formulation poses unique challenges: the model must anticipate forthcoming transitions while maintaining temporal consistency across abstraction levels. 

\section{PARSE: Hierarchical Partonomy Extraction}
\label{sec:approach}

We model the emergence of hierarchical partonomies as the natural outcome of predictive learning under uncertainty. 
Intuitively, an intelligent agent observing a video stream should continually anticipate future perceptual states, compare its predictions against reality, and reorganize its internal temporal representation when prediction errors exhibit sustained deviations. 
Such deviations correspond to event boundaries, i.e., moments where the predictive model must reconfigure its internal hypotheses. 
We model this as a hierarchical predictive learning framework where each predictive layer generates expectations about the layer below, receives prediction-error feedback, and updates its state only when error signals exceed adaptive thresholds.  
Trained under the streaming (causal) constraint, this hierarchy learns a continually evolving segmentation of time, a \emph{partonomy of experience}, that mirrors human perception of events as nested, temporally bounded processes.  
Figure~\ref{fig:arch} visualizes this mechanism, showing how hierarchical prediction, transient error dynamics, and the resulting nested event structure arise from a single unified objective. 

Let the incoming video be a sequence of frames $\{x_t\}_{t=1}^T$, with visual features $f_t = \phi(x_t)$ extracted by a visual feature encoder $\phi$. 
The predictive hierarchy consists of $N$ recurrent modules 
$\{\mathcal{F}^{(1)}, \mathcal{F}^{(2)}, \ldots, \mathcal{F}^{(N)}\}$, 
each maintaining an internal hidden state $h_t^{(i)}$ that evolves at its own temporal granularity. 
At every step, the system jointly minimizes the total predictive energy

\begin{equation}
\mathcal{E} = \sum_{i=1}^{N} \Big( \mathcal{L}_{\text{pred}}^{(i)} + \lambda_s\, \mathcal{L}_{\text{sparse}}^{(i)} \Big)
\label{eq:energy}
\end{equation}
where $\lambda_s$ balances the influence of sparsity regularization.
During streaming inference, the per-level prediction errors
$\mathcal{L}_{\text{pred}}^{(i)}(t)$
form temporal traces whose transient peaks signal structural changes in the input.
These error traces provide cues for hierarchical event segmentation and partonomy construction, avoiding the need for supervision or an explicit differentiable loss term. 

\subsection{Hierarchical Predictive Dynamics}
\label{subsec:hier_dyn}

Each layer in the predictive hierarchy operates as a recurrent coding unit that maintains both a \emph{local temporal hypothesis} and a \emph{contextual prior} from higher layers. 
The hierarchical dynamics are designed to instantiate a computational analogue of predictive coding with episodic feedback: each level anticipates the evolution of its subordinate representations while continually contextualizing them using the recent memory of higher-level states. 
This allows the model to infer nested temporal boundaries in a streaming fashion by implicitly tracking the evolving partonomy of ongoing experience. 
The lowest level, $L_1$, models fast, motion-driven transitions in the visual stream, i.e., those corresponding to short, motoric sub-events. 
Its hidden state evolves as a Markovian recurrence conditioned on both its previous representation and the incoming frame features:
\begin{equation}
h_{t+1}^{(1)} = 
\mathcal{F}^{(1)}(h_t^{(1)}, f_t),
\label{eq:l1_dynamics}
\end{equation}
where $\mathcal{F}^{(1)}$ is a recurrent predictor (e.g., RNN or LSTM~\cite{hochreiter1997long}) that learns local temporal dependencies. 
Higher levels evolve more slowly, integrating over extended horizons to form stable representations of mid- and high-level activities.
Rather than depending only on the immediate lower-layer state $h_t^{(i-1)}$, each level $L_i$ contextualizes its predictions using a trailing memory of the past $K$ hidden states from the layer above, $\{h_{t-k}^{(i+1)}\}_{k=1}^K$.
This design introduces a mechanism for top-down temporal contextualization, where higher-level hypotheses guide perception and segmentation at lower scales, analogous to goal-driven constrained event parsing in cognitive models of event perception~\cite{zacks2001event,tversky2008structure,tversky2014partonomies}.

Formally, the forward estimate produced by layer~$L_i$ for the next state of its subordinate layer~$L_{i-1}$ is given by:
\begin{equation}
\hat{h}_{t+1}^{(i-1)} =
\mathcal{F}^{(i)}\!\Big(
h_t^{(i-1)},\,
\mathrm{Attn}\big(h_t^{(i)}, \{h_{t-k}^{(i+1)}\}_{k=1}^K\big)
\Big)
\label{eq:hier_pred}
\end{equation}
where $i \in [2, N]$ and $\mathrm{Attn}(\cdot)$ denotes a scaled dot-product attention operator: $\mathrm{Attn}(q, M) =  \mathrm{softmax}\!\left( \frac{q M^\top}{\sqrt{d}} \right) M,$
where $q=h_t^{(i)}$, $M=\{h_{t-k}^{(i+1)}\}_{k=1}^K$, and $d$ the embedding dimension. 
This mechanism adaptively weights recent higher-level hypotheses, enabling each layer to interpret local changes in light of its inferred position within a broader temporal context. 
The level-wise prediction loss compares the predictor’s output to the appropriate next-step target-frame features at the bottom level and the lower-level hidden state otherwise. Let
\begin{equation}
s_{t+1}^{(i-1)} =
\begin{cases}
f_{t+1}, & i = 1,\\[3pt]
h_{t+1}^{(i-1)}, & i \ge 2,
\end{cases}
\qquad
\hat{s}_{t+1}^{(i-1)} = \mathcal{F}^{(i)}(\cdot).
\end{equation}
The prediction loss at level~$i$ is then defined as the normalized sum-squared error 
\begin{equation}
\mathcal{L}_{\text{pred}}^{(i)} =
\frac{1}{d_{i-1}}\,
\big\| \hat{s}_{t+1}^{(i-1)} - s_{t+1}^{(i-1)} \big\|_2^2,
\label{eq:pred}
\end{equation}
where $d_{i-1}$ denotes the dimensionality of $s_{t+1}^{(i-1)}$. 
Minimizing this hierarchical loss drives a dual process of inference and anticipation. 
Lower layers learn to predict short-term visual dynamics, while upper layers evolve more slowly, forming temporally abstract hypotheses that serve as contextual priors. 
The attention-mediated top-down flow provides adaptive feedback, comparable to \emph{partonomy consultation} in Event Segmentation Theory (EST)~\cite{zacks2001event,zacks2001perceiving,tversky2008structure,tversky2014partonomies}, allowing the model to re-evaluate its segmentation hypotheses based on longer-term structural expectations. 
This interaction between bottom-up evidence accumulation and top-down partonomic context enables the system to form temporally aligned, multi-scale representations, capturing immediate perceptual regularities and long-range event perception.

\textbf{Sparsity and Temporal Smoothness.} 
Although $\mathcal{L}{\text{pred}}$ encourages predictive accuracy, it can overfit to transient noise, producing unstable latent activations and overly frequent segmentation.
To regularize the hidden dynamics, we introduce an activation sparsity prior as $\mathcal{L}{\text{sparse}}^{(i)} = \big| h_{t}^{(i)} \big|_1.$ 
This $L_1$ penalty constrains the magnitude of hidden activations, encouraging compact, energy-efficient representations punctuated by sparse bursts of activity that mark significant state changes.
From an information-theoretic standpoint, it minimizes representational entropy, reducing spurious oscillations while retaining high sensitivity to meaningful transitions.
Empirically, this term stabilizes hierarchical training and ensures that higher layers evolve more slowly, capturing progressively coarser abstractions. 

\subsection{Generic Hierarchical Boundary Inference}

When a structural change occurs in the environment, such as the onset of a new action, the local prediction error $\mathcal{L}_{\text{pred}}^{(i)}(t)$ exhibits a sharp transient increase.
We exploit this property to detect event boundaries directly from the temporal dynamics of prediction errors, without introducing an explicit differentiable loss term.
Following streaming inference, each per-level error sequence is first smoothed using a short moving average (window size $K{=}3$) and then differentiated twice to accentuate inflection points.
Boundaries $\mathcal{B}^{(i)}$ are extracted as local maxima of the doubly differentiated, smoothed prediction-error signal within a neighborhood of radius $r_i$, formally:
\begin{equation}
\begin{split}
\mathcal{B}^{(i)} =
\Big\{\, t \;\Big|\;
&\Delta^2\!\big[\mathrm{MA}(\mathcal{L}_{\text{pred}}^{(i)}, K)\big]_t
>
\Delta^2\!\big[\mathrm{MA}(\mathcal{L}_{\text{pred}}^{(i)}, K)\big]_\tau,\\
&\forall\, \tau \in \mathcal{N}_{r_i}(t)
\Big\}
\end{split}
\label{eq:bound}
\end{equation}
where $\mathcal{N}_{r_i}(t)$ denotes a temporal neighborhood of radius $r_i$ (set by doing a grid search over the range of $0.1{\times}\mathrm{FPS}$ to $2{\times}\mathrm{FPS}$ for each level of the prediction stack). 
This procedure yields sparse, temporally coherent boundary sets $\{B^{(1)}, \ldots, B^{(N)}\}$ corresponding to hypothesized transitions at progressively coarser temporal scales.
Higher-level boundaries naturally emerge around sustained accumulations of prediction error, aligning with coarser temporal structure and producing nested event partitions.

\subsubsection{Emergent Partonomy Structure}

As the model processes streaming input, the latent hierarchy self-organizes into a multiscale representation of ongoing activity.
Each layer contributes a temporally segmented view of the sequence, and their alignment through containment induces a compositional structure:

\[
\mathcal{P} = \{ L_1, L_2, \ldots, L_N \}, \quad
L_i = \{ e_1^{(i)}, \ldots, e_{M_i}^{(i)} \},
\]
where each segment $e_j^{(i)}$ spans consecutive frames between adjacent boundaries in $B^{(i)}$.
Temporal containment between levels ensures that finer events aggregate consistently into coarser ones, forming a nested decomposition of the input stream. 
This emergent organization provides both interpretability and efficiency.
Instead of predicting discrete symbolic labels, the model learns to segment continuous experience into hierarchically aligned temporal units, allowing downstream reasoning modules,  e.g., affordance inference or plan recognition, to operate at an appropriate abstraction level.  
Because prediction, segmentation, and structural alignment are all consequences of minimizing the unified energy in Equation~\ref{eq:energy}, the system requires no external supervision or task-specific boundaries.

\textbf{Implementation Details.} 
We use a frozen EfficientNet-B2~\cite{tan2021efficientnetv2}, pretrained on ImageNet~\cite{deng2009imagenet}, as our primary choice of visual encoder, while also experiment with ResNet-50~\cite{he2016deep} and ViT/B32~\cite{dosovitskiyimage}. An LSTM~\cite{hochreiter1997long} is chosen as a recurrent predictor at each level, in a three-layer prediction stack. For top-down contextualization, each predictor attends at level $L_i$ over the last $K{=}5$ hidden states of the predictor at level $L_{i+1}$, maintained using a FIFO queue. The entire stack is trained simultaneously in a streaming manner over the training set of each dataset, once using the ADAM~\cite{adam2014method} optimizer with a learning rate of $1e^{-3}$. During inference, the learning rate is dropped to $1e^{-6}$ to allow the model to adapt to changes within a video, but is reset to the trained weights between videos to prevent learning transfer across videos. $\lambda_{bound}$ and $\lambda_{sparse}$ are set to $1.0$ and $0.1$, respectively. The boundary predictions are obtained by applying a moving-average filter set to 20\% of the video frame rate and identifying transient peaks within a sliding window of size $(0.25\times \mathrm{FPS},\, 1.25\times \mathrm{FPS},\, 2.00\times \mathrm{FPS})$ for the three hierarchical levels (tuned per dataset via grid search). All models are trained on a workstation server with an AMD Threadripper 64-core processor, 512 GB RAM, and 4 NVIDIA RTX 6000 ADA, one video at a time for streaming training. More details in the supplementary.

\begin{table}[t]
  \centering
  \small
  \setlength{\tabcolsep}{1pt}
  \begin{tabular}{l l c c cc}
    \toprule
    \textbf{Dataset} & \textbf{Model} & \textbf{Streaming} & \textbf{Oracle} & \textbf{TED} & \textbf{hF1} \\
    \midrule
    \multirow{6}{2cm}{\textbf{Breakfast Actions}}
      & Fixed Length    & \ding{51} & \ding{51} & \textbf{38.42} & 21.46 \\
      & K-Means          & \ding{55} & \ding{51} & 16.18 & 8.60 \\
      & Oracle K-Means   & \ding{55} & \ding{51}* & 16.58 & 9.27 \\
      & Hierarchical    & \ding{55} & \ding{51} & 33.35 & \underline{22.60} \\ \cmidrule{2-6}
       & Streamer        & \ding{51} & \ding{55} & 18.72 & 8.68 \\
      & {PARSE (Ours)} & \ding{51} & \ding{55} & \underline{36.04} & \textbf{26.81} \\
    \midrule
    \multirow{6}{*}{\textbf{50 Salads}}
      & {Fixed Length} & \ding{51} & \ding{51} & \textbf{54.33} & \textbf{25.73} \\
      & K-Means          & \ding{55} & \ding{51} & 6.87  & 2.70 \\
      & Oracle K-Means   & \ding{55} & \ding{51}* & 6.85  & 2.71 \\
      & Hierarchical    & \ding{55} & \ding{51} & \underline{35.59} & \underline{17.98} \\ \cmidrule{2-6}
      & Streamer        & \ding{51} & \ding{55} & 14.34 & 3.70 \\
      & PARSE (Ours)          & \ding{51} & \ding{55} & 20.65 & 10.76 \\
    \midrule
    \multirow{6}{2cm}{\textbf{Assembly 101}}
      & Fixed Length    & \ding{51} & \ding{51} & 27.66 & {21.10} \\
      & K-Means          & \ding{55} & \ding{51} & 10.85 & 5.42 \\
      & Oracle K-Means   & \ding{55} & \ding{51}* & 11.00 & 5.57 \\
      & Hierarchical    & \ding{55} & \ding{51} & \underline{30.20} & \underline{21.76} \\ \cmidrule{2-6}
      & Streamer        & \ding{51} & \ding{55} & 15.04 & 8.83 \\
      & {PARSE (ours)} & \ding{51} & \ding{55} & \textbf{32.77} & \textbf{21.99} \\
    \bottomrule
  \end{tabular}
  \caption{Partonomy inference performance (\%). 
  ``Streaming'' indicates whether the method operates in a streaming (causal) setting. 
  ``Oracle'' denotes whether the approach assumes access to dataset-specific priors, such as the average number of segments to expect or the average length of a segment. * indicates ground truth usage. 
  }
  \label{tab:partonomy_all}
\end{table}

\begin{table*}[t]
  \centering
  \small
  \setlength{\tabcolsep}{3.5pt}
  \begin{tabular}{l l c c *{4}{c} *{4}{c}}
    \toprule
    \textbf{Dataset} & \textbf{Model} & \textbf{Streaming} & \textbf{Oracle} &
    \multicolumn{4}{c}{\textbf{Fine-grained}} &
    \multicolumn{4}{c}{\textbf{Coarse-grained}} \\
    \cmidrule(lr){5-8} \cmidrule(lr){9-12}
    & & & & \textbf{Precision} & \textbf{Recall} & \textbf{mIoU} & \textbf{F1} & \textbf{Precision} & \textbf{Recall} & \textbf{mIoU} & \textbf{F1} \\
    \midrule
    \multirow{6}{*}{\textbf{Breakfast Actions}}
      & Fixed Length    & \ding{51} & \ding{51} & 22.11 & 26.56 & 13.44 & 24.13 & 8.95 & 25.78 & 6.45 & 13.29 \\
      & K-means         & \ding{55} & \ding{51} & 14.23 & \textbf{66.67} & 12.97 & 23.45 & 4.24 & \textbf{75.88} & 4.14 & 8.03 \\
      & Oracle K-means  & \ding{55} & \ding{51}* & 14.39 & 64.59 & 12.97 & 23.54 & 4.30 & 75.70 & 4.20 & 8.14 \\
      & Hierarchical    & \ding{55} & \ding{51} & 21.61 & 43.04 & 16.36 & 28.77 & 9.82 & 48.85 & 8.62 & 16.35 \\ \cmidrule{2-12}
      & Streamer        & \ding{51} & \ding{55} & 14.17 & 40.90 & 11.13 & 21.05 & 5.50 & 42.49 & 4.18 & 9.74 \\
      & {PARSE (ours)} & \ding{51} & \ding{55} & \textbf{22.48} & 58.57 & \textbf{18.93} & \textbf{32.49} & \textbf{13.61} & 34.28 & \textbf{10.98} & \textbf{19.48} \\
    \midrule
    \multirow{6}{*}{\textbf{50 Salads}}
      & Fixed Length    & \ding{51} & \ding{51} & \textbf{6.29} & 5.77 & 3.10 & 6.02 & 3.80 & 5.57 & 2.31 & 4.52 \\
      & K-means         & \ding{55} & \ding{51} & 2.68 & 41.60 & 2.57 & 5.04 & 1.69 & 57.23 & 1.67 & 3.28 \\
      & Oracle K-means  & \ding{55} & \ding{51}* & 2.75 & 42.19 & 2.64 & 5.16 & 1.75 & \textbf{60.39} & 1.73 & 3.40 \\
      & Hierarchical    & \ding{55} & \ding{51} & 5.42 & 13.09 & 4.00 & 7.67 & 4.65 & 21.19 & 3.97 & 7.63 \\ \cmidrule{2-12}
      & Streamer        & \ding{51} & \ding{55} & 4.77 & 34.70 & 4.35 & 8.39 & 1.50 & 25.90 & 1.44 & 2.84 \\
      & {PARSE (ours)} & \ding{51} & \ding{55} & 5.66 & \textbf{52.05} & \textbf{5.10} & \textbf{10.21} & \textbf{5.19} & 24.53 & \textbf{4.40} & \textbf{8.57} \\
    \midrule
    \multirow{6}{*}{\textbf{Assembly 101}}
      & Fixed Length    & \ding{51} & \ding{51} & 20.07 & 22.27 & 11.57 & 21.11 & 5.81 & 21.99 & 4.31 & 9.19 \\
      & K-means         & \ding{55} & \ding{51} & 12.27 & \textbf{70.41} & 11.60 & 20.90 & 1.77 & \textbf{79.10} & 1.76 & 3.46 \\
      & Oracle K-means  & \ding{55} & \ding{51}* & 12.39 & 69.82 & 11.68 & 21.05 & 1.78 & 78.42 & 1.77 & 3.48 \\
      & Hierarchical    & \ding{55} & \ding{51} & \textbf{26.54} & 48.09 & \textbf{20.42} & \textbf{34.20} & 6.46 & 40.71 & \textbf{5.78} & \textbf{11.15} \\ \cmidrule{2-12}
      & Streamer        & \ding{51} & \ding{55} & 16.24 & 30.12 & 11.71 & 21.10 & 1.98 & 26.69 & 1.88 & 3.69 \\
      & {PARSE (ours)} & \ding{51} & \ding{55} & 20.55 & 63.92 & 18.34 & 31.10 & \textbf{7.25} & 18.66 & 5.34 & 10.44 \\
    \bottomrule
  \end{tabular}
    \caption{Hierarchical GEBD performance. Fine-grained and Coarse-grained boundary metrics. ``Streaming'' indicates whether the method operates causally. 
    PARSE consistently outperforms other streaming baselines or offers competitive performance to non-streaming ones.
    }
  \label{tab:fine_coarse_all}
\end{table*}
\section{Experimental Evaluation}\label{sec:eval}
\textbf{Datasets.}
We evaluate \textit{PARSE} on three benchmark datasets with multi-scale temporal annotations:
(1) \textit{Breakfast Actions}~\cite{kuehne2014language}, 1,712 third-person videos of 10 breakfast routines annotated at action (fine) and activity (coarse) levels. We use 325 videos for training and 48 with dual-level labels for testing.
(2) \textit{50 Salads}~\cite{stein2013combining}, over 4 hours of egocentric videos of 25 participants preparing two salads each, with fine-grained (e.g., cut tomato) and coarse-grained (e.g., prepare ingredients) labels. We train on 30 and test on 20 clips.
(3) \textit{Assembly101}~\cite{sener2022assembly101}, 4,321 egocentric videos of toy-vehicle assembly and disassembly with frame-level fine/coarse annotations. We use 114 clips for training and 200 for evaluation.
For all datasets, two-level ground-truth partonomies are constructed by nesting fine-grained segments within their corresponding coarse events, ensuring temporal containment and structural coherence (Eq.~\ref{eq:containment}). The supplementary shows qualitative visualizations for deeper partonomies extracted by PARSE. 

\textbf{Tasks and Metrics.}
We assess both temporal and structural fidelity through two complementary tasks:
(i) \textit{Hierarchical Generic Event Boundary Detection (H-GEBD)} measures how well prediction-error transients align with annotated boundaries across scales. Following prior GEBD protocols, we report precision, recall, and mean-IoU within a temporal tolerance window.
(ii) \textit{Partonomy Structure Prediction} evaluates hierarchical consistency using Tree Edit Distance (TED) and hierarchical F1 (hF1). TED quantifies the cost of transforming the predicted hierarchy into the ground truth, while hF1 measures temporal alignment of corresponding segments across levels. Together, these metrics reflect the objectives in Eq.~\ref{eq:energy}, where $\mathcal{L}{\text{pred}}$ governs temporal precision and $\mathcal{L}{\text{sparse}}$ and $\mathcal{L}_{\text{bound}}$ regularize hierarchical structure. Full definitions are in the Supplementary.

\textbf{Baselines.}
We compare against streaming and offline methods differing in temporal granularity and structural priors.
\textit{Fixed Length} divides each video into uniform-duration segments based on dataset-level averages.
\textit{K-means} clusters frame-level features into the average number of events per hierarchy level, while \textit{K-means (Oracle)} uses ground-truth event counts per video as an upper bound.
\textit{Hierarchical Linkage} forms a temporal dendrogram over frame embeddings and extracts two-level partonomies by cutting at dataset-average event counts.
\textit{STREAMER}~\cite{mounir2023streamer} is a transformer-based predictive baseline that trains layers sequentially, adding cross-layer communication after each stage to output boundary predictions per scale.
\textit{PARSE} is our proposed, unsupervised hierarchical predictive learner that operates fully online, jointly optimizing all layers under streaming constraints. 
All baselines use the same visual encoder (EfficientNetv2) for fairness.

\begin{figure}[t]
    \centering
    \includegraphics[width=\columnwidth]{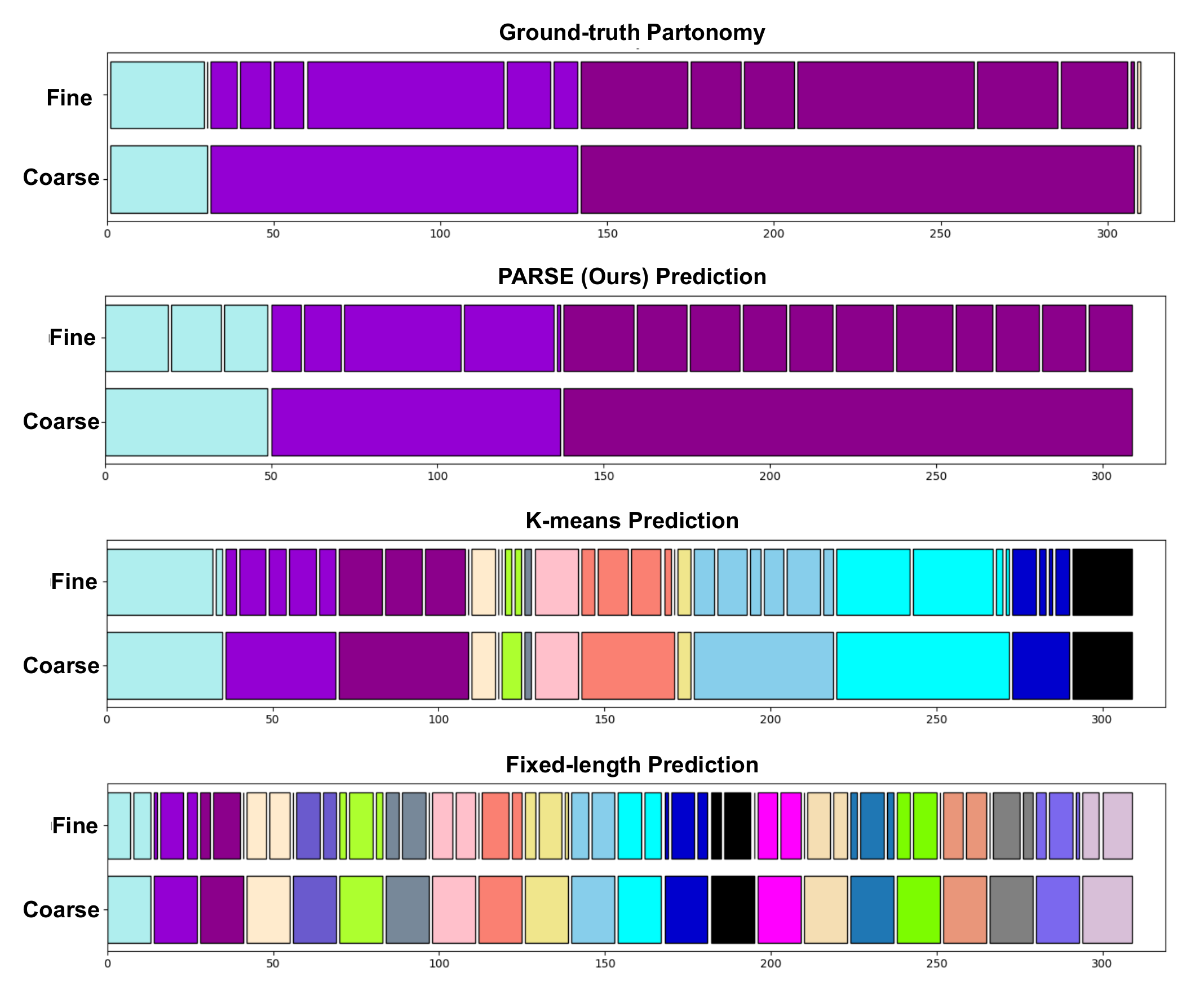}     \\    
    \caption{\textbf{Qualitative comparison of hierarchical partonomy predictions.}
  We visualize the two best-matching levels (fine and coarse) from each predicted hierarchy for a video ``Making coffee'' from Breakfast Actions. 
  PARSE produces temporally coherent and hierarchically consistent segments.
  }
  \label{fig:qualitative_partonomy}
\end{figure}

\subsection{Event Partonomy Prediction Performance}

Table \ref{tab:partonomy_all} reports quantitative results for the partonomy prediction task using the Tree Edit Distance (TED) and hierarchical F1 (hF1) metrics. These measures capture complementary properties of hierarchical segmentation: TED prioritizes structural integrity, i.e., how well predicted parent–child relations and containment are preserved, while hF1 reflects temporal alignment and consistency across abstraction levels. Because these objectives often trade off, higher values in one metric do not always correspond to superior segmentation quality. 
Across datasets, \textit{PARSE} achieves the best overall balance of TED and hF1 among streaming (causal) methods, consistently outperforming \textit{STREAMER} and other non-oracle baselines. The \textit{Fixed Length} baseline attains competitive TED scores on highly regular routines such as \textit{Breakfast Actions} and \textit{50 Salads}, where activity durations are approximately uniform; however, it fails under variable-length settings like \textit{Assembly101}, where strict periodic segmentation violates temporal alignment. Clustering-based approaches (\textit{K-means} and its oracle variant) and the \textit{Hierarchical Linkage} method perform segmentation retrospectively: they require access to the entire video, rely on dataset- or ground-truth-derived heuristics (e.g., average event count or length), and construct partonomies post hoc. As a result, they cannot adapt to evolving uncertainty or streaming evidence. In contrast, \textit{PARSE} operates fully online, discovering boundaries and adjusting hierarchical structure on the fly. 

\begin{figure*}[t]
  \centering
  \begin{subfigure}[t]{0.32\textwidth}
    \includegraphics[width=\linewidth]{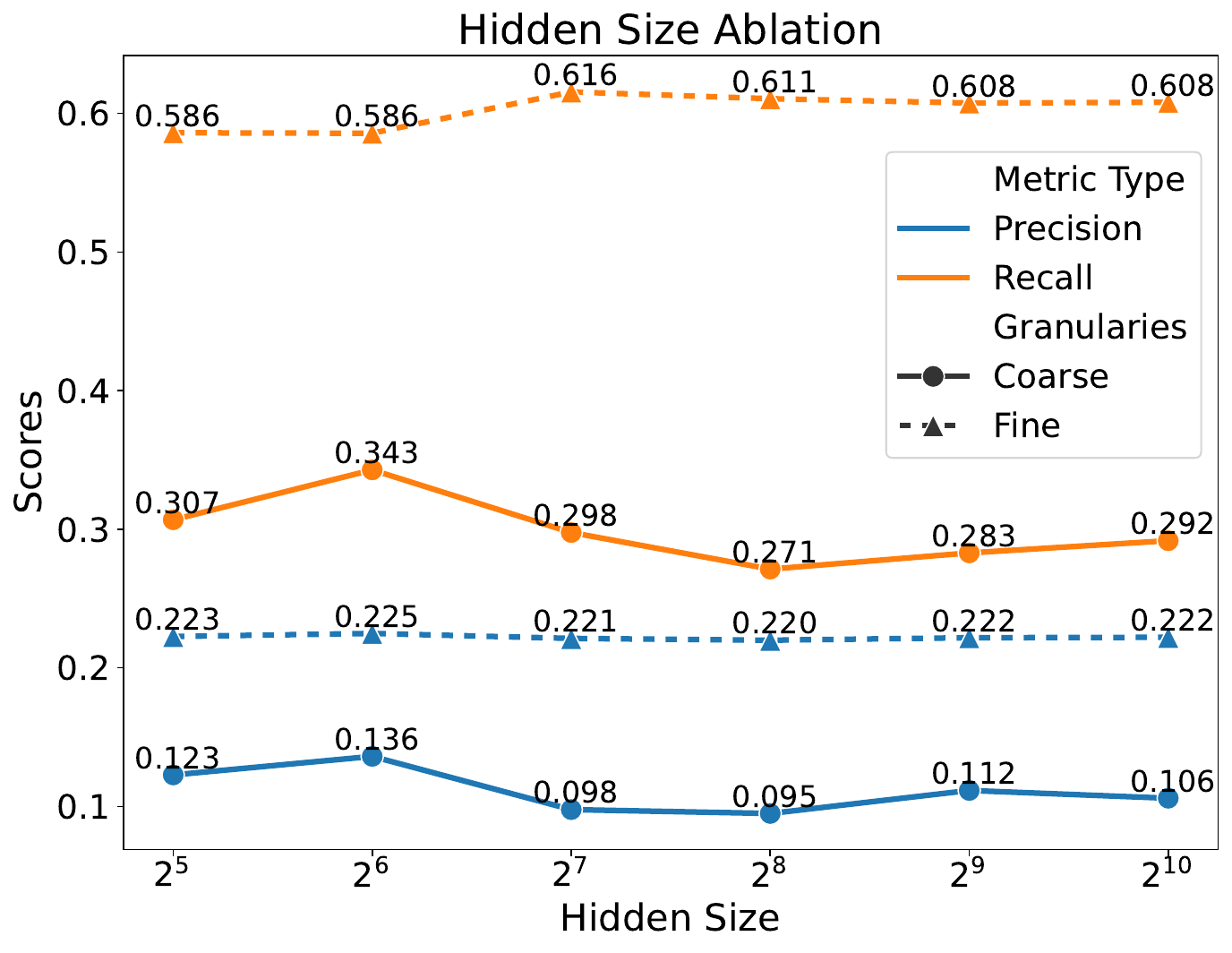}
    \caption{Hidden state dimension}
    \label{fig:ablation_hidden}
  \end{subfigure}
  \begin{subfigure}[t]{0.32\textwidth}
    \includegraphics[width=\linewidth]{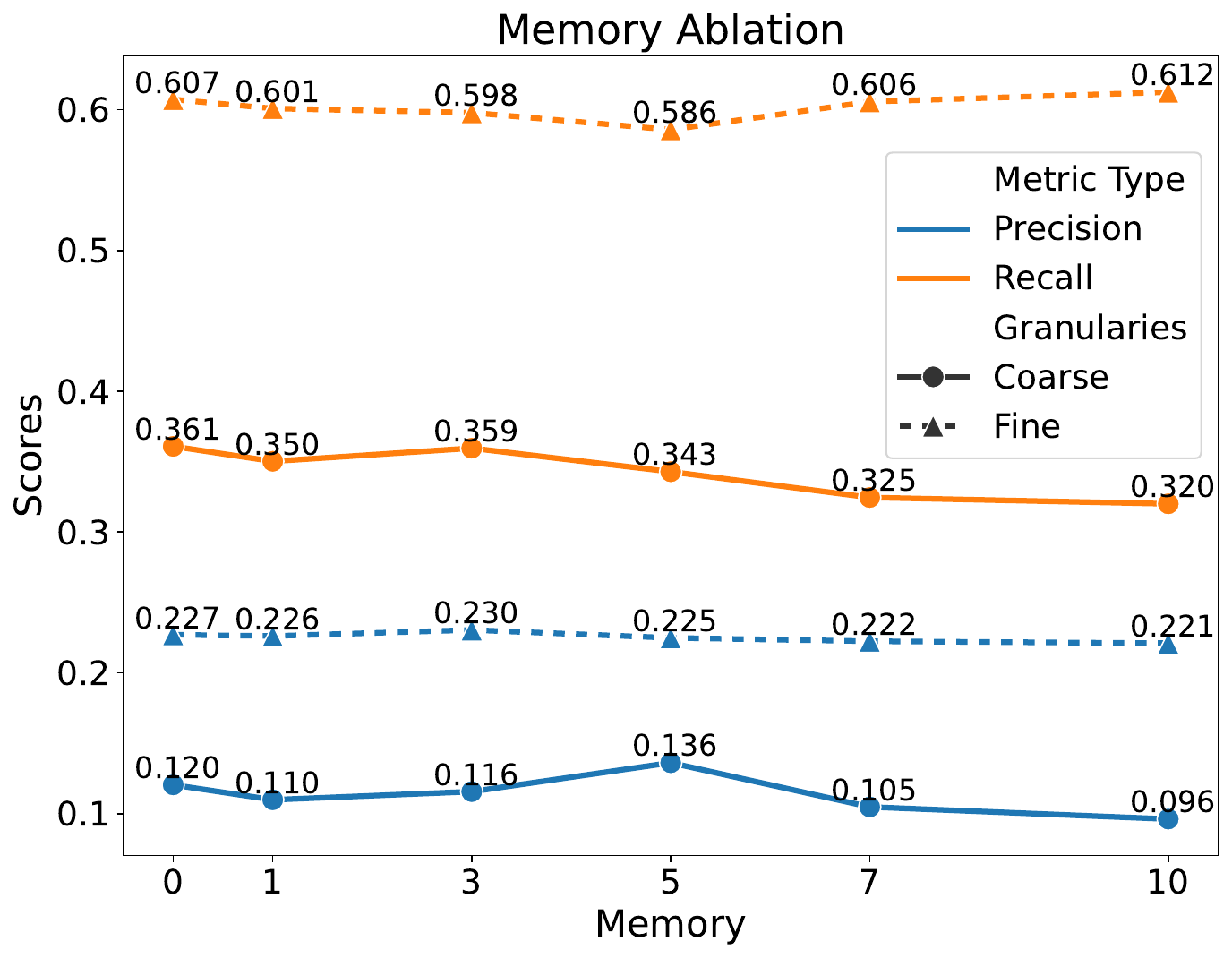}
    \caption{Memory window $K$}
    \label{fig:ablation_memory}
  \end{subfigure}
  \begin{subfigure}[t]{0.32\textwidth}
    \includegraphics[width=\linewidth]{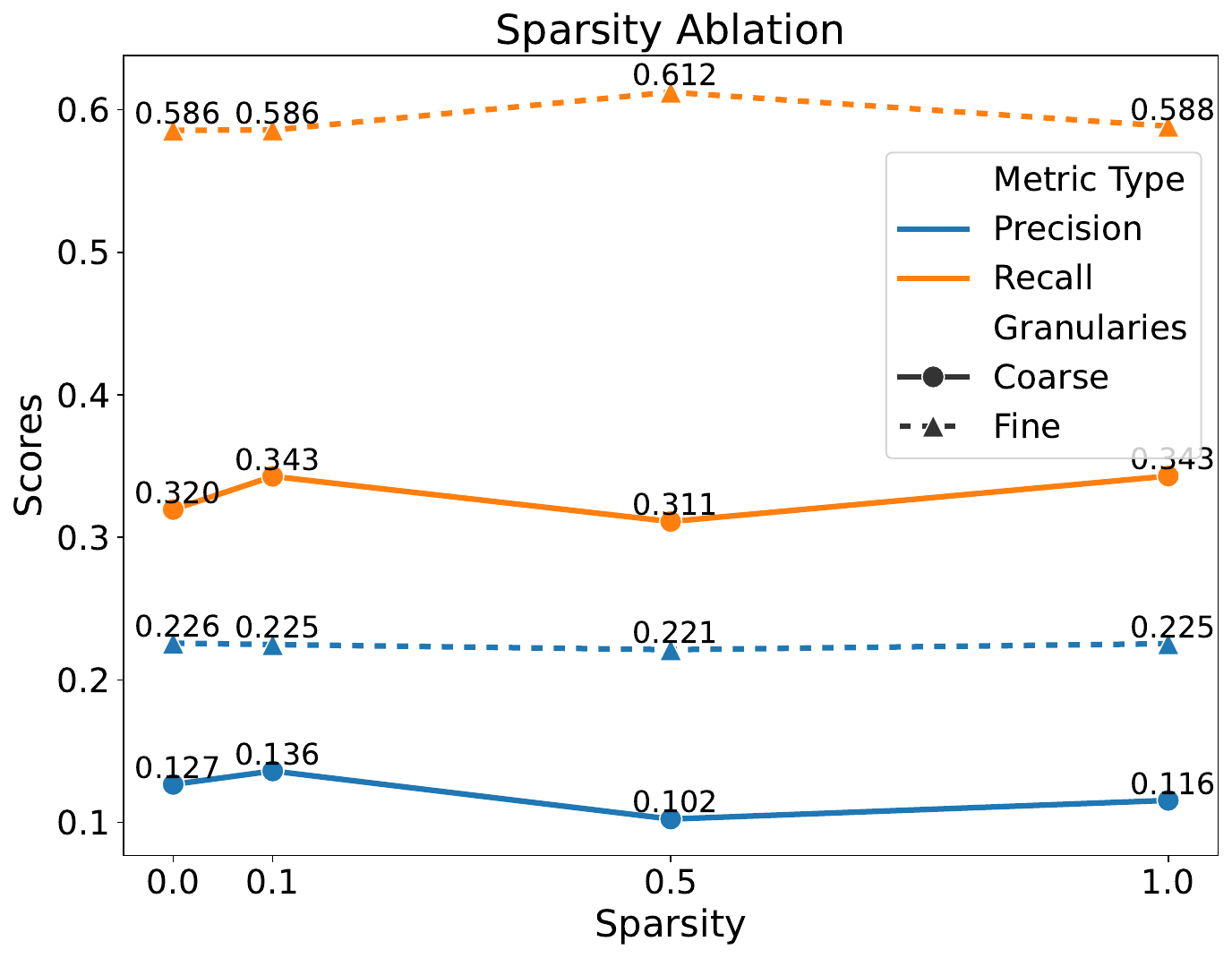}
    \caption{Sparsity weight $\lambda_s$}
    \label{fig:ablation_sparsity}
  \end{subfigure}
  \caption{\textbf{Ablation on dynamic hyperparameters.}
  Effects of hidden-state size, top-down memory length, and sparsity regularization on boundary precision and recall at fine and coarse scales. Moderate configurations yield the most stable predictive dynamics.}
  \label{fig:ablation_dynamic}
\end{figure*}

\subsection{Hierarchical GEBD Performance}
Table~\ref{tab:fine_coarse_all} reports the fine- and coarse-grained boundary detection results for all methods. This task emphasizes temporal segmentation fidelity rather than hierarchical structure, with precision and recall measuring selectivity and coverage of predicted transitions, and mIoU/F1 summarizing their alignment quality. In contrast to the structural TED–hF1 metrics in Table~\ref{tab:partonomy_all}, these boundary-level measures are sensitive to over-segmentation and boundary drift, revealing the temporal stability of each approach. 
Across all datasets, \textit{PARSE} achieves the highest or near-highest F1 and mIoU among streaming (causal) methods, demonstrating strong temporal coherence despite operating without access to future frames. The advantage is most evident on complex sequences such as \textit{Breakfast Actions} and \textit{Assembly101}, where boundary distributions are irregular and event durations vary widely. By coupling predictive anticipation with temporal smoothing, \textit{PARSE} avoids the excessive fragmentation seen in \textit{STREAMER} and the missed transitions characteristic of static segmentation methods. 
Offline baselines (\textit{K-means}, \textit{Hierarchical Linkage}) appear to achieve high recall or precision on certain datasets, but these results stem from retrospective access to the entire video and heuristic assumptions about event counts or durations. 
In contrast, \textit{PARSE} infers boundaries directly from streaming prediction-error dynamics, adapting online to event variability and yielding consistent fine and coarse performance. 

\begin{table}[t]
  \centering
  \setlength{\tabcolsep}{5pt}
  \begin{tabular}{lcccc}
    \toprule
    \textbf{Configuration} & \textbf{Fine F1} & \textbf{Coarse F1} & \textbf{TED} & \textbf{hF1} \\
    \midrule
    \multicolumn{5}{c}{\textit{Predictor Model}}\\
    \midrule
    LSTM & \textbf{32.49} & \textbf{19.48} & \textbf{36.04} & \textbf{26.81}\\
    Transformer & 32.31 & 18.12 & 34.85 & 24.34\\
    \midrule
    \multicolumn{5}{c}{\textit{Visual Encoder}}\\
    \midrule
    
    EfficientNet-B2 & \textbf{32.49} & \textbf{19.48} & \textbf{36.04} & \textbf{26.81}\\
    ResNet-50       & 32.37 & 18.99 & 35.75 & 26.53\\
    ViT/B-32        & 32.32 & 13.97 & 34.76 & 24.49 \\
    \midrule
    \multicolumn{5}{c}{\textit{Number of Prediction Layers}}\\
    \midrule
    2 Layers        & 32.34 & 18.46 & 34.96 & 24.16 \\
    3 Layers        & 32.49 & \textbf{19.48} & \textbf{36.04} & \textbf{26.81}\\
    4 Layers        & 32.75 & 16.39 & 34.38 & 23.11\\
    5 Layers        & \textbf{33.26} & 18.82 & 35.24 & 24.25\\
    \bottomrule
  \end{tabular}
\caption{Architectural ablations on the type of predictor at each layer, the visual backbone, and prediction hierarchy depth. Metrics are reported on the Breakfast Actions Dataset.}
  \label{tab:arch_ablation}
  
\end{table}

\textbf{Qualitative Discussion}
Figure~\ref{fig:qualitative_partonomy} illustrates representative two-level partonomies for the ground truth and different models. \textit{PARSE} closely replicates the hierarchical organization of real activities, producing temporally coherent segments whose fine-level events remain properly contained within their corresponding coarse events. In contrast, most baselines, particularly \textit{Fixed Length} and \textit{K-means}, violate this containment constraint, as their recursive hierarchy construction often produces misaligned or fragmented boundaries across levels. These violations manifest as color discontinuities and inconsistent coarse–fine segmentation, revealing that static or clustering-based approaches cannot maintain temporal nesting when operating independently at each level. \textit{PARSE}, by contrast, infers boundaries directly from hierarchical prediction-error dynamics, yielding partonomies that adapt naturally to event duration. 

\subsection{Ablation Studies}


\textbf{Architectural choices.} As shown in Table~\ref{tab:arch_ablation}, PARSE is robust across predictor types and visual backbones. 
Replacing the LSTM predictor with a Transformer yields similar performance but slightly weaker coarse-scale alignment, suggesting that recurrent gating better preserves temporal continuity under streaming constraints. 
Among encoders, EfficientNet-B2 achieves the highest overall consistency (TED = 36.0, hF1 = 26.8), while ResNet-50 and ViT/B-32 perform comparably despite differing in spatial bias. 
Varying the number of predictive layers reveals that three layers offer the best trade-off between temporal precision and structural depth: deeper hierarchies show minor gains in fine-grained recall but reduced hierarchical coherence. 

\textbf{Dynamic hyperparameters.} Figure~\ref{fig:ablation_dynamic} examines the effect of sparsity weight~$\lambda_s$, hidden-state dimension~$d_h$, and memory window length~$K$ on both boundary and structural metrics. 
Moderate sparsity ($\lambda_s{=}0.1$) consistently yields the highest coarse-level and partonomy scores.
Lower values cause dense, unstable activations, whereas larger weights suppress adaptive changes, leading to under-segmentation. 
Hidden-state size shows a similar trend: increasing $d_h$ up to 64 improves representational capacity, but further enlargement reduces coarse-level consistency due to oversmoothing. 
Memory window $K{=}5$ performs best, balancing short-term responsiveness with long-range contextualization. 
\section{Discussion, Limitations, and Future Work} \label{sec:conclusion}
This paper introduced PARSE, a unified hierarchical predictive framework that learns multiscale event structure directly from streaming video without supervision or dataset-specific priors. By coupling predictive anticipation with sparsity-driven temporal regularization, PARSE infers coherent partonomies whose fine- and coarse-level boundaries emerge naturally from the dynamics of prediction error. Quantitative and qualitative evaluations across three datasets demonstrate that the model achieves competitive structural and temporal alignment while operating entirely online.
While promising, our formulation remains limited by the granularity and reliability of existing hierarchical annotations and by the scale of available training data. In the future, we aim to extend PARSE to richer datasets, continuous long-horizon streams, and more expressive (and deeper) event structures to better capture compositional structure in real-world tasks for embodied event perception. 

\textbf{Acknowledgments.} This work was supported in part by the US NSF grants IIS 2348689 and IIS 2348690, and the USDA award no. 2023-69014-39716. The authors would like to thank the authors of the Breakfast Actions, Assembly-101, and 50-Salads for making their data and annotations publicly available. 

{
    \small
    \bibliographystyle{ieeenat_fullname}
    \bibliography{main}
}

\newpage

\section*{Appendix A}

\begin{figure*}[t]
    \centering
    \begin{tabular}{cc}
         \multicolumn{2}{c}{\textbf{Groundtruth 3-level Partonomy}}\\
        \includegraphics[width=0.45\textwidth]{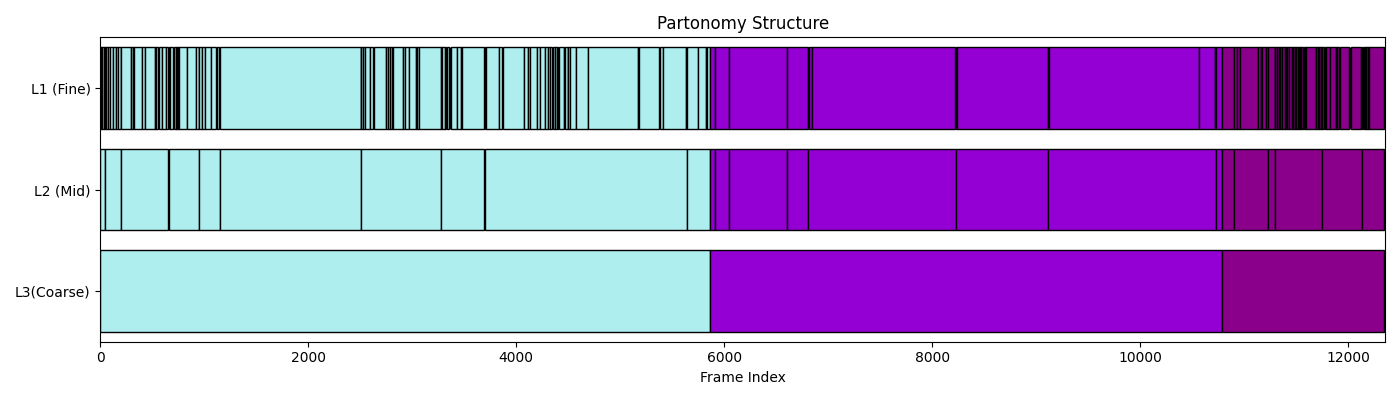} & 
        \includegraphics[width=0.45\textwidth]{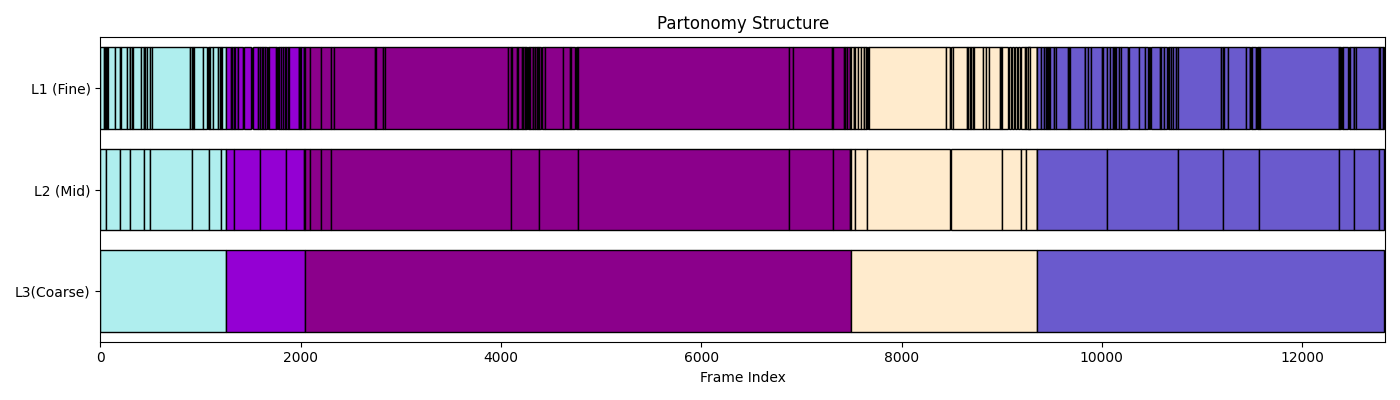}\\
         \multicolumn{2}{c}{\textbf{Predicted 3-level Partonomy}}\\
         \includegraphics[width=0.45\textwidth]{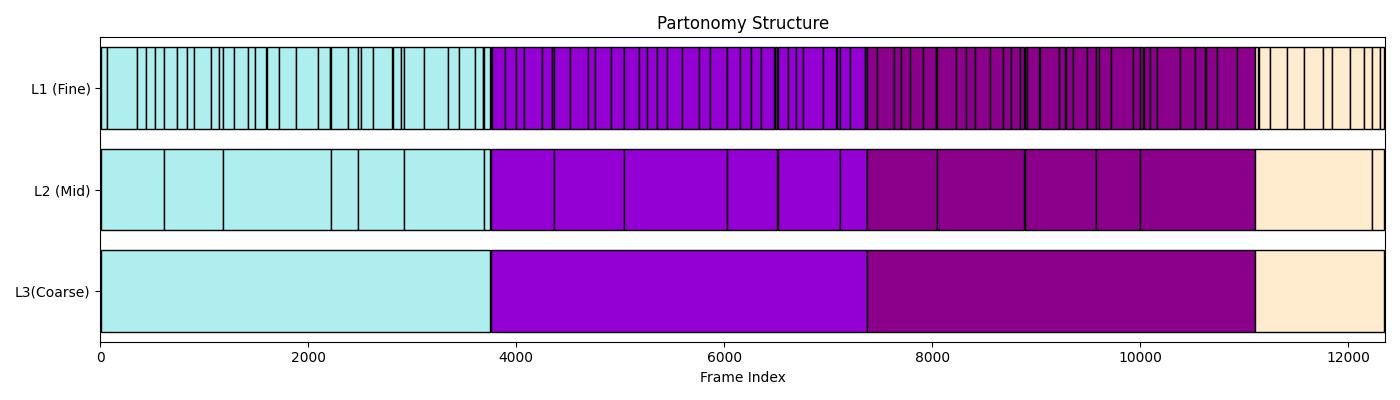} & 
        \includegraphics[width=0.45\textwidth]{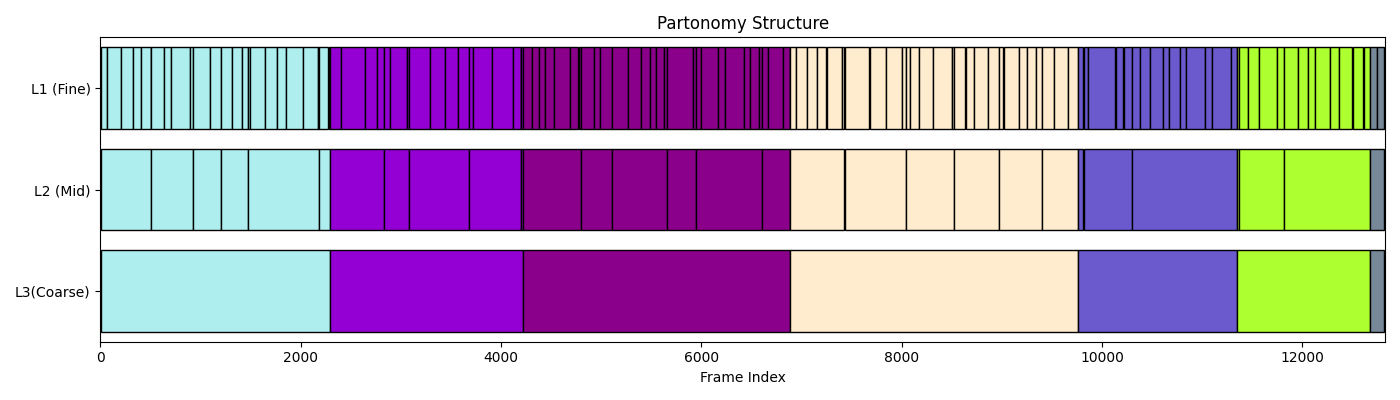}\\
    \end{tabular}
    \caption{\textbf{Three-level partonomy predictions} for simulated long videos for subjects P44 and P48 from the Breakfast Actions dataset.}
    \label{fig:threelevel_results}
\end{figure*}
\section{Evaluation Metrics and Protocols}
\label{sec:supp_metrics}

We evaluate our model using two complementary tasks designed to measure both
(\textit{i})~the temporal accuracy of detected boundaries across abstraction levels and
(\textit{ii})~the structural correctness of the resulting partonomy.
These correspond to the behavioral manifestations of the optimization objective in
Equation 2: 
the predictive consistency term ($\mathcal{L}_{\text{pred}}$) aligns with temporal precision and recall,
while the sparsity ($\mathcal{L}_{\text{sparse}}$) and boundary regularization ($\mathcal{L}_{\text{bound}}$) 
shape the emergence of stable, hierarchically aligned structures.
Together, they form the computational substrate for the two evaluation tasks described below.

\subsection{Hierarchical Generic Event Boundary Detection (H-GEBD)}

The first task assesses \emph{temporal segmentation accuracy} at each level~$L_i$ of the predicted hierarchy.
Let the predicted boundary set be $\mathcal{B}_p^{(i)} = \{b_1, \ldots, b_{M_p}\}$ and
the ground-truth boundary set be $\mathcal{B}_g^{(i)} = \{g_1, \ldots, g_{M_g}\}$.
A predicted boundary~$b_k$ is considered correct if it lies within a tolerance window of $\pm w$ frames around any ground-truth boundary:
\begin{equation}
\delta(b_k, \mathcal{B}_g^{(i)}) =
\begin{cases}
1, & \exists \, g_j \in \mathcal{B}_g^{(i)} \text{ s.t. } |b_k - g_j| \le w, \\
0, & \text{otherwise.}
\end{cases}
\end{equation}
Using this criterion, we compute:
\begin{align}
TP = \sum_k \delta(b_k, \mathcal{B}_g^{(i)}) \\
FP = M_p - TP, \quad
FN = M_g - TP,
\end{align}
and define the standard metrics:
\begin{align}
P = \frac{TP}{TP + FP} \\
R = \frac{TP}{TP + FN} \\
\text{mIoU} = \frac{TP}{TP + FP + FN}.
\label{eq:hgebd_metrics}
\end{align}
Precision~($P$) measures the selectivity of boundary detections, 
recall~($R$) quantifies the completeness of the segmentation, and
mIoU measures the overlap consistency between predicted and ground-truth boundaries.
Across levels, these scores reveal how the model’s hierarchical prediction errors 
translate into temporally aligned boundary transitions, thereby validating the effect of $\mathcal{L}_{\text{pred}}$ and $\mathcal{L}_{\text{bound}}$ in Equation 2.

\subsection{Partonomy Structure Prediction}

While H-GEBD evaluates boundary timing, it does not assess whether the predicted boundaries yield a coherent hierarchical structure. 
We therefore evaluate the emergent partonomy
$\hat{\mathcal{P}} = \{\hat{L}_1, \hat{L}_2, \ldots, \hat{L}_N\}$
against the ground truth 
$\mathcal{P}^* = \{L_1^*, L_2^*, \ldots, L_N^*\}$
using two complementary structural metrics:
the \emph{Tree Edit Distance (TED)} and the \emph{Hierarchical F1-score (hF1)}.

\paragraph{Tree Edit Distance (TED).}
TED measures \textit{structural correctness} by quantifying the minimal cost of transforming
$\hat{\mathcal{P}}$ into $\mathcal{P}^*$ through insertions, deletions, or relabelings.
Following~\cite{zhang1989simple}, let each node correspond to an event segment $e_j^{(i)}$,
and define the elementary edit operations:
\begin{align}
\text{cost}(\text{insert}) = \alpha \\
\text{cost}(\text{delete}) = \alpha \\
\text{cost}(\text{relabel}) = \beta \, d_\text{label}(e_a, e_b),
\end{align}
where $d_\text{label}$ is $1$ for label mismatch and $0$ otherwise. Since our task is unsupervised, we always set $\beta=0$ and $\alpha=1$
The TED between the two partonomies is then defined recursively as
\begin{equation}
\text{TED}(\hat{\mathcal{P}}, \mathcal{P}^*) =
\min_{\pi \in \Pi} 
\sum_{(a,b) \in \pi}
\text{cost}(a \!\rightarrow\! b),
\label{eq:ted}
\end{equation}
where $\Pi$ is the set of all valid edit paths between corresponding trees.
We normalize the TED by the total number of nodes to obtain a similarity score in $[0,1]$:
\begin{equation}
\text{TED-Sim} = 1 - 
\frac{\text{TED}(\hat{\mathcal{P}}, \mathcal{P}^*)}
     {\alpha (|\hat{\mathcal{P}}| + |\mathcal{P}^*|)}.
\label{eq:ted_sim}
\end{equation}
A higher TED-Sim indicates better preservation of hierarchical structure and parent–child containment as defined in Equation 1.

\paragraph{Hierarchical F1 (hF1).}
While TED evaluates discrete structure, hF1 measures \textit{temporal consistency} across abstraction levels.
At each depth~$i$, we first extract the set of segments 
$L_i = \{e_1^{(i)}, \ldots, e_{M_i}^{(i)}\}$
and $\hat{L}_i = \{\hat{e}_1^{(i)}, \ldots, \hat{e}_{\hat{M}_i}^{(i)}\}$.
For each pair $(e_a^{(i)}, \hat{e}_b^{(i)})$, we compute their intersection-over-union:
\begin{equation}
\text{IoU}(e_a^{(i)}, \hat{e}_b^{(i)}) = 
\frac{|e_a^{(i)} \cap \hat{e}_b^{(i)}|}
     {|e_a^{(i)} \cup \hat{e}_b^{(i)}|}.
\end{equation}
We perform bipartite matching via the Hungarian algorithm to maximize total IoU,
and count a match if $\text{IoU} \ge \tau$.
The level-wise F1-score is then:
\begin{align}
F_1^{(i)} = 
\frac{2 P^{(i)} R^{(i)}}
     {P^{(i)} + R^{(i)}} \\
P^{(i)} = \frac{TP^{(i)}}{TP^{(i)} + FP^{(i)}}, \;
R^{(i)} = \frac{TP^{(i)}}{TP^{(i)} + FN^{(i)}}.
\label{eq:hf1}
\end{align}
The overall hierarchical F1 is the mean across levels:
\begin{equation}
\text{hF1} = \frac{1}{N} \sum_{i=1}^{N} F_1^{(i)}.
\end{equation}
High hF1 values indicate that predicted event boundaries not only align temporally with ground-truth segments but also respect the containment and ordering constraints across scales.

\subsection{Interpretation}

TED and hF1 together provide a complete picture of partonomy inference performance.
TED reflects the \emph{structural well-formedness} of the predicted hierarchy---how faithfully the model reconstructs the parent–child relationships between nested events.
hF1 reflects the \emph{temporal containment fidelity}---whether segment boundaries at different scales remain hierarchically consistent.
These metrics complement the H-GEBD measures from Eq.~\ref{eq:hgebd_metrics}:
while boundary-level precision and recall are sensitive to the temporal localization of change,
TED and hF1 are sensitive to the \textit{structural coherence} that emerges through minimizing the energy objective in Equation 2.
Together, they evaluate whether the model not only predicts \textit{when} event transitions occur but also \textit{how} these transitions compose to form a temporally coherent, multiscale partonomy.

\begin{figure*}
    \centering
    \begin{tabular}{cc}
    \midrule
    \textbf{Assembly-101} & \textbf{50 Salads} \\
    \midrule
    \multicolumn{2}{c}{\textbf{Ground truth}}\\
    \midrule
      \includegraphics[width=0.45\textwidth]{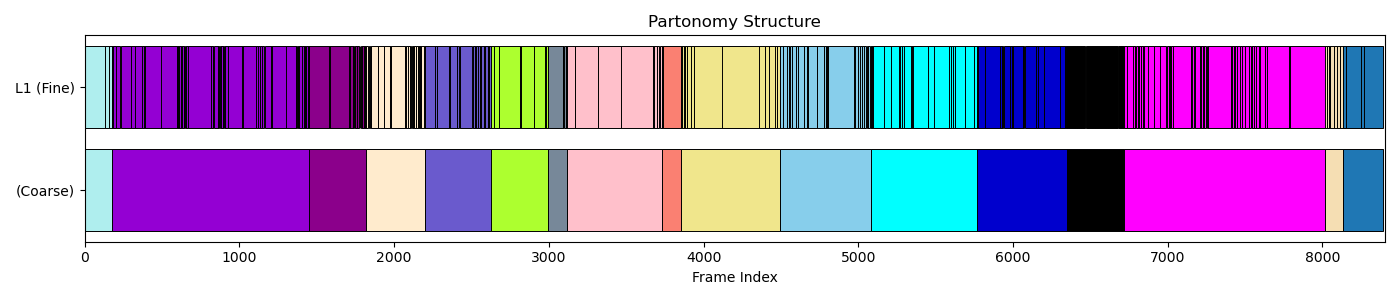} & \includegraphics[width=0.45\textwidth]{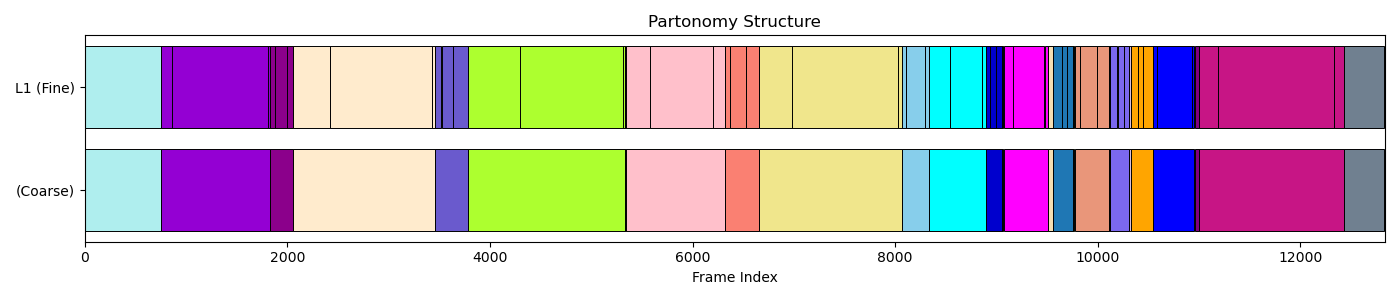} \\
      \midrule
    \multicolumn{2}{c}{\textbf{PARSE Predictions}}\\
    \midrule
      \includegraphics[width=0.45\textwidth]{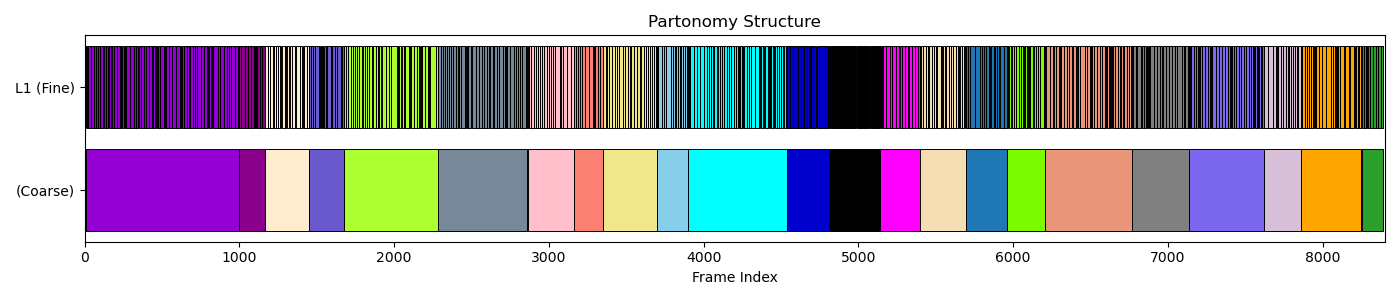} & \includegraphics[width=0.45\textwidth]{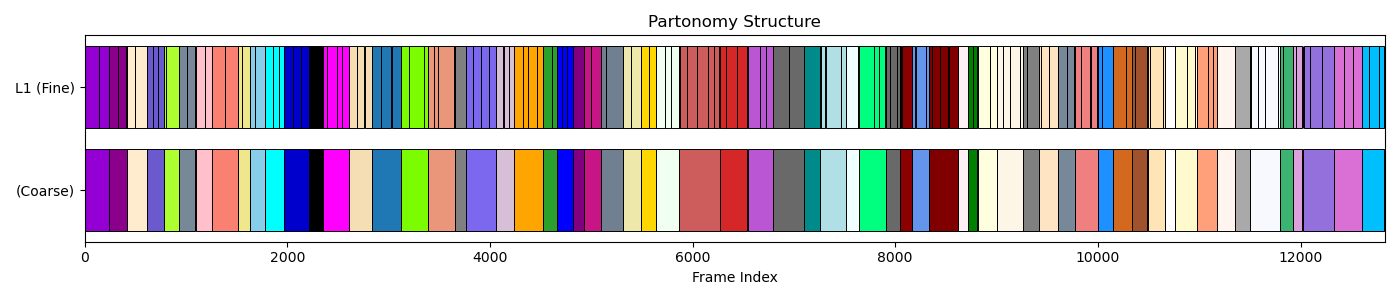} \\
      \midrule
    \multicolumn{2}{c}{\textbf{STREAMER Predictions}}\\
    \midrule
      \includegraphics[width=0.45\textwidth]{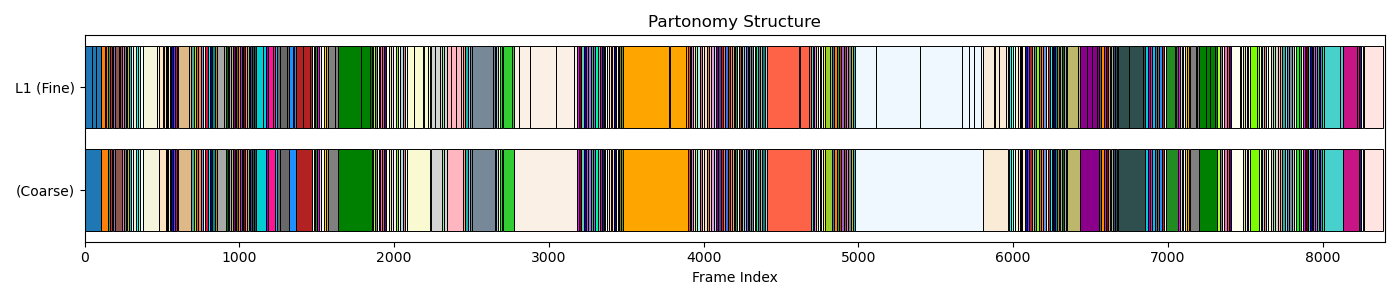} & \includegraphics[width=0.45\textwidth]{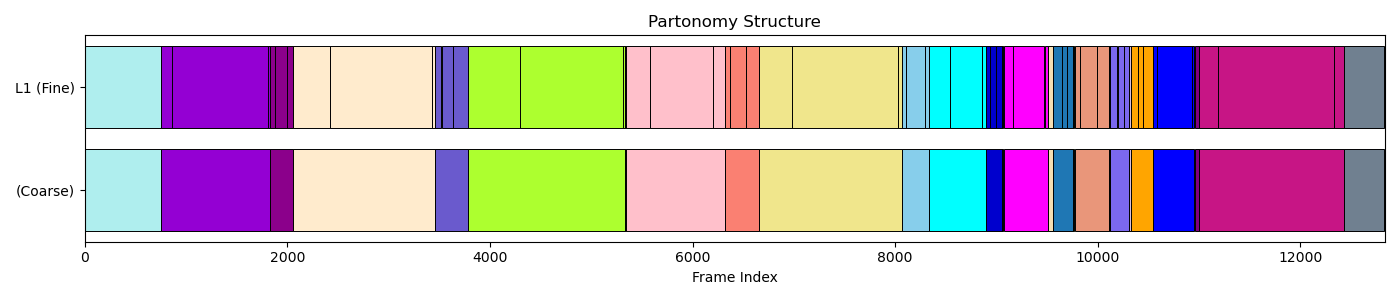} \\
      \midrule
    \multicolumn{2}{c}{\textbf{K-Means Predictions}}\\
    \midrule
      \includegraphics[width=0.45\textwidth]{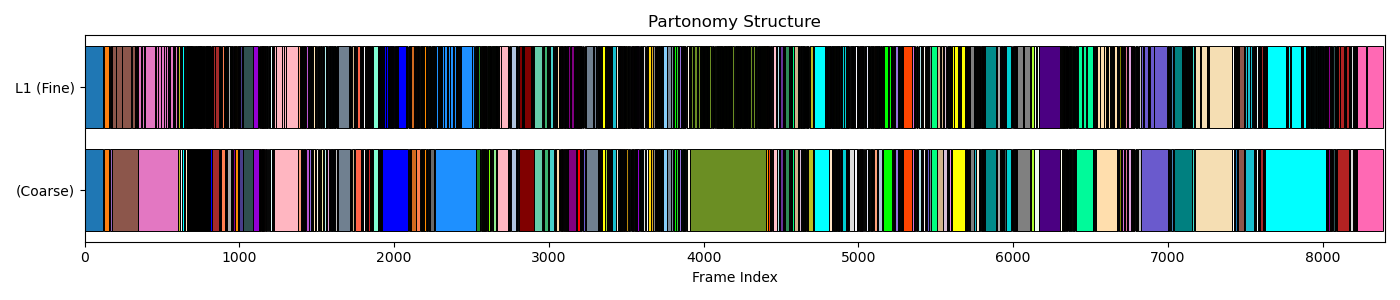} & \includegraphics[width=0.45\textwidth]{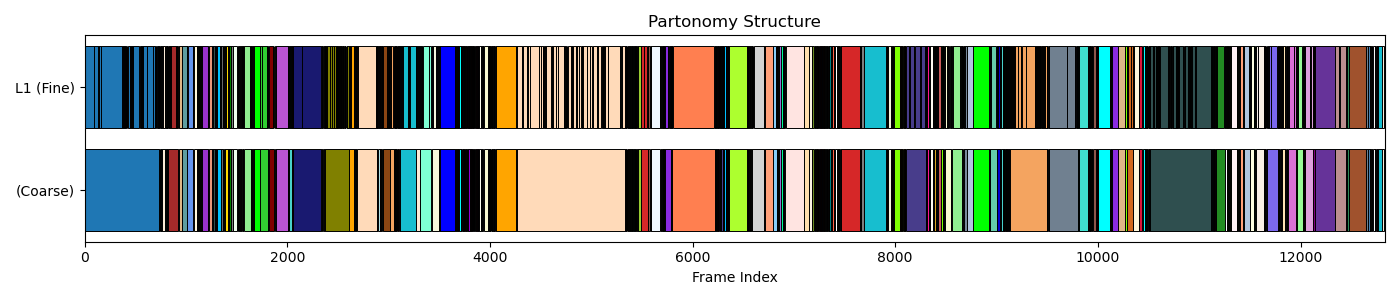} \\

    \midrule
    \multicolumn{2}{c}{\textbf{Hierarchical Predictions}}\\
    \midrule
      \includegraphics[width=0.45\textwidth]{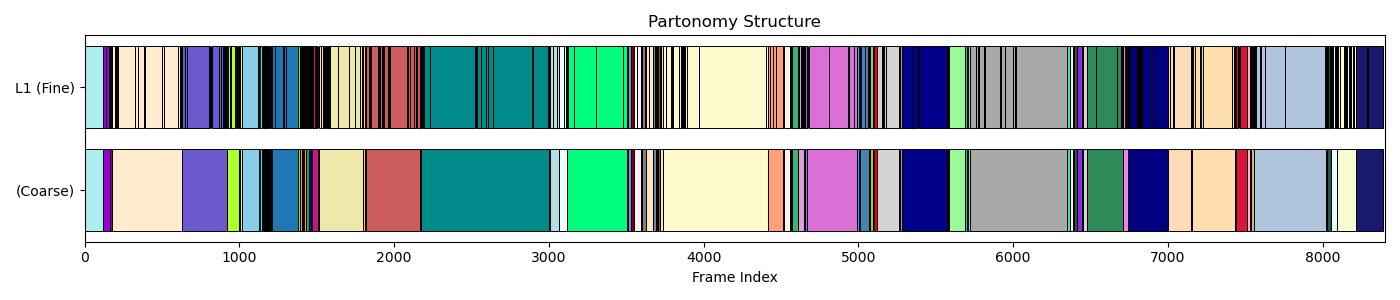} & \includegraphics[width=0.45\textwidth]{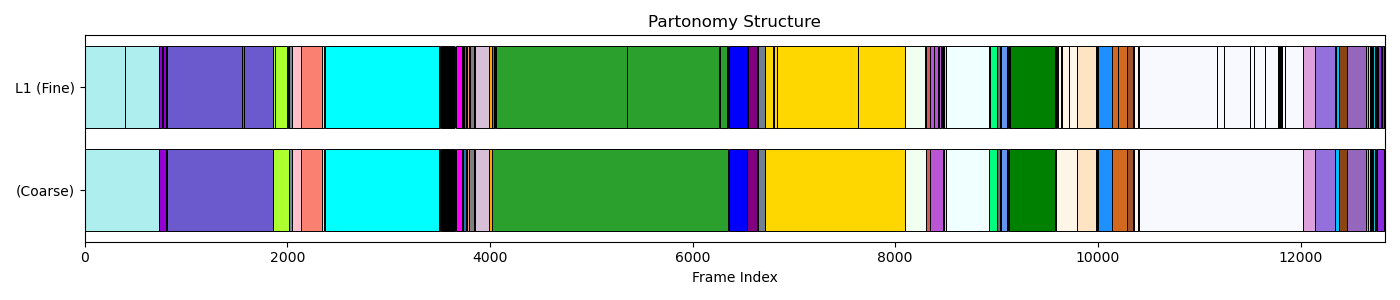} \\
    \end{tabular}
    \caption{\textbf{Additional Qualitative Visualizations} of the predictions from different baselines on the Assembly-101 (left) and 50 Salads (right) datasets.}
    \label{fig:new_qual}
\end{figure*}

\section{Detailed Experimental Setup}
\textbf{Data.} We repurpose three publicly available datasets with multi-scale temporal annotations to evaluate our model on two tasks: (i) partonomy inference, which measures the recovery of nested event hierarchies, and (ii) hierarchical generic event boundary detection (H-GEBD), which assesses temporal segmentation consistency across abstraction levels.
Breakfast Actions \cite{kuehne2014language} contains 1,712 third-person videos of 10 breakfast preparation activities performed by 52 actors. Each video is annotated at both fine-grained (action) and coarse-grained (activity) levels, enabling two-level hierarchical evaluation. We use 48 videos with dual-level annotations for testing and sample 325 for training.
50 Salads \cite{stein2013combining} includes over 4 hours of egocentric recordings of 25 participants each preparing two mixed salads. Videos are annotated with both low-level action segments (e.g., cut tomato) and high-level activity labels (e.g., prepare ingredients). We evaluate on 20 videos containing both levels of annotation and train on 30 additional clips.
Assembly101 \cite{sener2022assembly101} is a large-scale egocentric dataset comprising 4,321 videos of participants assembling and disassembling 101 toy vehicles with frame-level fine and coarse annotations. We sample 114 clips for training and 200 for evaluation.
For all datasets, we construct two-level partonomies by recursively nesting fine-grained temporal segments within their corresponding coarse-grained segments. Empirically, we find that reliable ground-truth partonomies can be constructed up to two levels without violating temporal containment or structural coherence assumptions defined in Equation 1.

\textbf{Evaluation Tasks and Metrics.} We evaluate our model on two complementary tasks that jointly assess the temporal and structural fidelity of inferred partonomies. The first task, Hierarchical Generic Event Boundary Detection (H-GEBD), measures how well transient prediction-error peaks align with annotated event transitions at multiple temporal scales. Following prior GEBD formulations, we compute precision, recall, and mean-IoU between predicted and ground-truth boundary frames within a tolerance window, quantifying both selectivity and coverage of temporal change. The second task, Partonomy Structure Prediction, evaluates the hierarchical consistency of the inferred event tree using the Tree Edit Distance (TED) and the hierarchical F1-score (hF1). TED captures the structural cost of transforming the predicted hierarchy into the ground truth, reflecting parent–child containment and compositional coherence, while hF1 measures the temporal alignment of corresponding segments across levels using IoU-based matching. Together, these metrics directly evaluate the objectives encoded in Equation 2: $\mathcal{L}_{\text{pred}}$ governs temporal precision and $\mathcal{L}_{\text{sparse}}$ and $\mathcal{L}_{\text{bound}}$ regulate hierarchical consistency. Full mathematical definitions of TED, hF1, and the boundary metrics are provided in the Supplementary.

\textbf{Baselines.} We compare our approach (\textit{PARSE}) against a diverse set of streaming and offline segmentation baselines that differ in how they infer multi-scale temporal structure.
\textit{Fixed Length} constructs two-level partonomies by partitioning each video into segments of uniform duration, where the average segment length at each hierarchy level is estimated from dataset statistics.
\textit{K-means} uses the average number of events per level in each dataset as the target cluster count, performing feature clustering on frame-level visual embeddings (identical to those used in our model) and recursively nesting the resulting segments into a two-level hierarchy.
\textit{K-means (Oracle)} follows the same procedure but sets the cluster count to the ground-truth number of events in each video, providing an upper bound on unsupervised clustering performance.
\textit{Hierarchical Linkage} constructs a fully connected temporal linkage matrix over frame embeddings and derives a two-level partonomy by cutting the dendrogram at levels corresponding to the average number of events per hierarchy level in the dataset.
\textit{STREAMER}~\cite{mounir2023streamer} is a transformer-based predictive model conceptually related to ours: it trains layers sequentially in a bottom-up fashion, training the first layer for 10k steps, then incrementally adding higher layers with cross-layer communication for an additional 10k steps each, to produce event boundary predictions at multiple temporal scales. The final partonomy is then constructed recursively as with the other baselines.
\textit{PARSE} provides a unified, end-to-end hierarchical predictive learner that jointly optimizes all layers under streaming constraints, producing temporally aligned and coherent partonomies without supervision or dataset-specific priors.

\section{Beyond 2-layer Partonomy}

This section provides a qualitative demonstration that \textit{PARSE} extends naturally beyond the two-level hierarchies used in our main experiments. Although the Breakfast Actions dataset provides only fine and coarse annotations, several participants perform multiple distinct tasks across separate videos. We use this structure to approximate a third partonomy level corresponding to high-level goals.

\noindent\textbf{Three-level hierarchical test set.}
For each participant, we identify all video instances that share the same participant ID and contain both fine and coarse annotations. These videos are concatenated in chronological order to create a single continuous sequence. The fine and coarse annotations associated with each original clip are also concatenated along the temporal dimension so that they remain aligned with the merged sequence. 
We then introduce a third annotation level, denoted L3, which records the boundaries between the original sub-videos in the concatenated sequence. Each L3 segment corresponds to one high-level task performed by that participant, for example preparing juice, preparing milk, or making a sandwich. The resulting hierarchy is:
\begin{itemize}
    \item L1: fine-grained action segments such as reaching for an object or pouring ingredients,
    \item L2: mid-level activity segments composed of multiple actions, such as preparing milk or assembling a sandwich,
    \item L3: goal-level segments corresponding to each original video associated with the participant.
\end{itemize}

\textbf{Qualitative results.}
Figure~\ref{fig:threelevel_results} shows ground truth and predicted three-level partonomies for one representative participant. Although this test scenario is synthetic and does not represent a naturally continuous recording, \textit{PARSE} produces a coherent nested hierarchy across all three levels. L1 boundaries remain dense and responsive to short-term changes, L2 boundaries reflect coherent multi-action activities, and L3 boundaries align with the larger goal transitions induced by video concatenation. Even in this setting, the predicted segments naturally respect containment across levels, demonstrating that the hierarchical consistency encouraged by the predictive cascade generalizes beyond the two-level structures used in training.

\textbf{Limitations.} 
We do not provide a quantitative evaluation for this experiment because concatenated videos do not form a continuous real-world activity stream. Camera resets and scene changes at clip boundaries violate the assumptions underlying our partonomy definition. Nevertheless, these qualitative results illustrate that the predictive hierarchy can be extended to deeper structures without modifications to the architecture or additional supervision.
\section{Qualitative Visualization}
Additional qualitative visualizations are shown in Figure~\ref{fig:new_qual} on Assembly-101 and 50 Salads datasets. As can be seen, PARSE consistently produces stable, nested segments across datasets.

\section{Detailed Implementation Details}
\subsection{Breakfast Actions}
\textbf{Hyperparameters for PARSE}

\medskip
\noindent\textbf{Model Configuration}
\begin{itemize}
    \item LSTM hidden size: 64
    \item LSTM layers: 1
\end{itemize}

\noindent\textbf{Training Configuration}
\begin{itemize}
    \item Learning rate: $1 \times 10^{-3}$
    \item Sparsity weight: 0.1
\end{itemize}

\noindent\textbf{Inference Configuration}
\begin{itemize}
    \item Learning rate: $1 \times 10^{-6}$
    \item Sparsity weight: 0.1
    \item Sliding history windows: $\{7,\,15,\,30\}$ for L2/L3/L4
\end{itemize}

\noindent\textbf{Postprocessing}
\begin{itemize}
    \item Moving average smoothing:
    $$
    \text{smooth}_{L2}=4,\quad 
    \text{smooth}_{L3}=3,\quad
    \text{smooth}_{L4}=2
    $$
    \item Transient peak orders:
    $$
    \text{order}_{L2}=4,\quad
    \text{order}_{L3}=20,\quad
    \text{order}_{L4}=30
    $$
\end{itemize}

\noindent\textbf{Implementation Details for Baselines:} 

\textbf{Setting $k$ for different granularities:}  
We parse the ground-truth annotations and remove non-informative events such as \texttt{SIL}.  
The average number of valid segments is computed separately for the fine- and coarse-level annotations.  
For the mid-level granularity, we define its segment count as the rounded average of these two values:
\[
k_{\mathrm{mid}} = \left\lceil \frac{k_{\mathrm{coarse}} + k_{\mathrm{fine}}}{2} \right\rceil .
\]
Using this procedure, the final values employed in all clustering-based baselines are:
\[
k_{\mathrm{coarse}} = 6, \quad
k_{\mathrm{mid}} = 22, \quad
k_{\mathrm{fine}} = 38 .
\]

\subsection{Assembly 101}

\textbf{Hyperparameters for PARSE:}

\medskip
\noindent\textbf{Model Configuration}
\begin{itemize}
    \item LSTM hidden size: 128
    \item LSTM layers: 1
\end{itemize}

\noindent\textbf{Training Configuration}
\begin{itemize}
    \item Learning rate: $1 \times 10^{-3}$
    \item Sparsity weight: 0.1
\end{itemize}

\noindent\textbf{Inference Configuration}
\begin{itemize}
    \item Learning rate: $1 \times 10^{-5}$
    \item Sparsity weight: 0.1
    \item Sliding history windows: $\{7,\,15,\,30\}$ for L2/L3/L4
\end{itemize}

\noindent\textbf{Postprocessing}
\begin{itemize}
    \item Moving average smoothing:
    $$
    \text{smooth}_{L2}=3,\quad 
    \text{smooth}_{L3}=3,\quad
    \text{smooth}_{L4}=3
    $$
    \item Transient peak orders:
    $$
    \text{order}_{L2}=4,\quad
    \text{order}_{L3}=30,\quad
    \text{order}_{L4}=60
    $$
\end{itemize}

\textbf{Implementation Details for Baselines.}

We follow the same procedure described in the Breakfast dataset. For this dataset, the resulting values are:
\[
k_{\mathrm{coarse}} = 26,\quad
k_{\mathrm{mid}} = 134,\quad
k_{\mathrm{fine}} = 242.
\]

\subsection{50 Salads}

\textbf{Hyperparameters for PARSE:}

\medskip
\noindent\textbf{Model Configuration}
\begin{itemize}
    \item LSTM hidden size: 512
    \item LSTM layers: 1
\end{itemize}

\noindent\textbf{Training Configuration}
\begin{itemize}
    \item Learning rate: $1 \times 10^{-3}$
    \item Sparsity weight: 0.1
\end{itemize}

\noindent\textbf{Inference Configuration}
\begin{itemize}
    \item Learning rate: $1 \times 10^{-6}$
    \item Sparsity weight: 0.1
    \item Sliding history windows: $\{7,\,15,\,30\}$ for L2/L3/L4
\end{itemize}

\noindent\textbf{Postprocessing}
\begin{itemize}
    \item Moving average smoothing:
    $$
    \text{smooth}_{L2}=3,\quad 
    \text{smooth}_{L3}=3,\quad
    \text{smooth}_{L4}=3
    $$
    \item Transient peak orders:
    $$
    \text{order}_{L2}=4,\quad
    \text{order}_{L3}=20,\quad
    \text{order}_{L4}=45
    $$
\end{itemize}

\textbf{Implementation Details for Baselines:}

For this dataset, the values are:
\[
k_{\mathrm{coarse}} = 19,\quad k_{\mathrm{mid}} = 36,\quad k_{\mathrm{fine}} = 52.
\]

\subsection{Impact of Order on Performance}
\begin{figure}[t]
    \centering
    \includegraphics[width=0.85\linewidth]{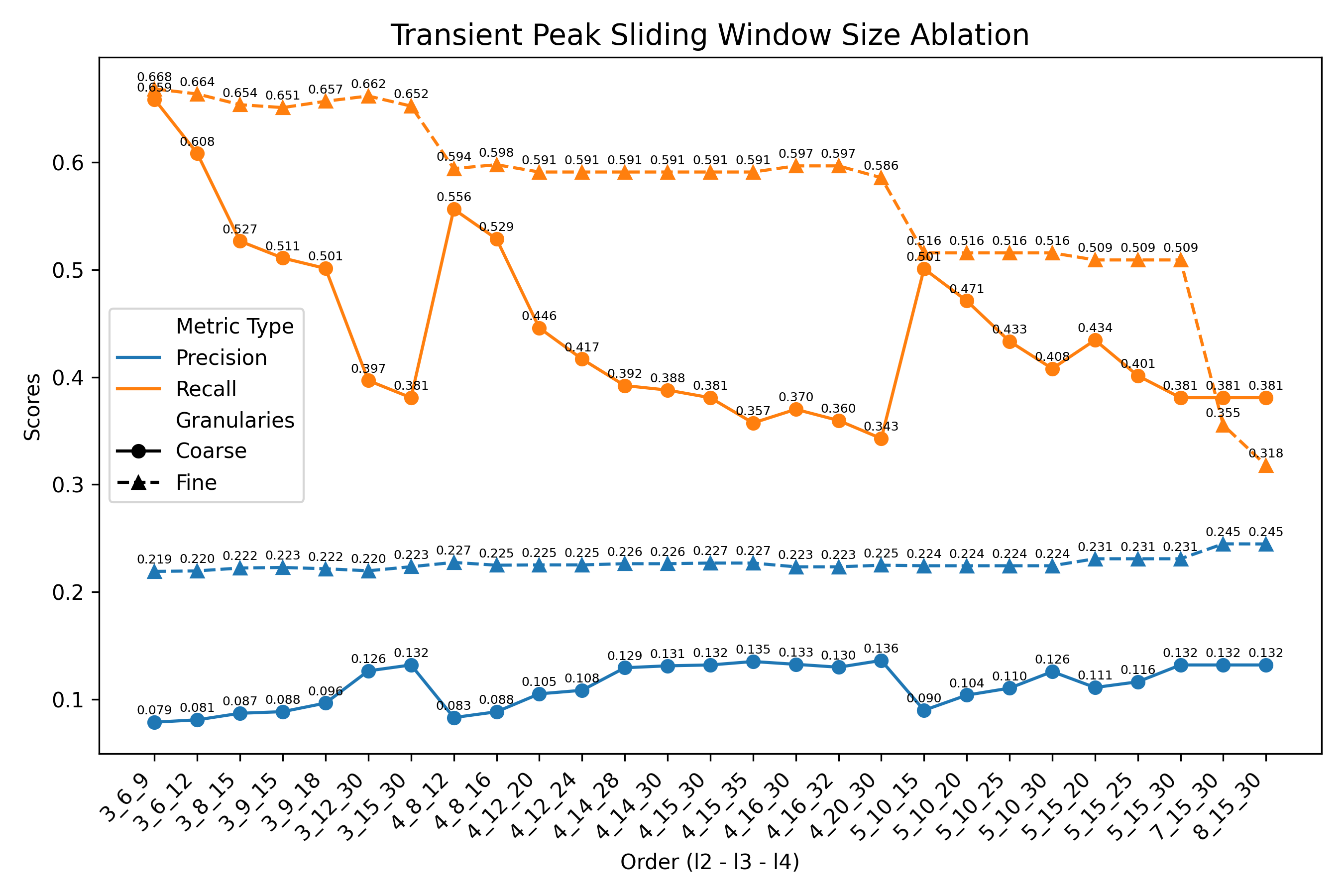}
    \caption{Ablation of the L2–L3–L4 order parameters on fine- and coarse-grained boundary detection, showing the precision–recall trade-off as the peak-detection neighborhood increases.}
    \label{fig:order_ablation}
\end{figure}

As shown in Figure~\ref{fig:order_ablation}, as the order parameter increases at the L2 level, fine-grained precision gradually improves while fine-grained recall drops sharply. In the context of argrelextrema, a larger order value requires each boundary candidate to be a strict local maximum over a broader neighborhood. This makes the detector more conservative at the fine scale, suppressing many short-lived or weak peaks that would otherwise be considered boundaries.

A similar trend is observed when increasing the L3 and L4 orders: coarse-level precision consistently increases, whereas coarse-level recall steadily decreases. Since boundary events at these levels tend to be longer and smoother, requiring a larger neighborhood for peak validation further reduces the number of detected boundaries.

Overall, increasing the order parameter enforces stricter peak validation by expanding the neighborhood
within which a point must dominate to be considered a boundary. This reduces the total number of predicted
boundaries, resulting in higher precision (fewer false positives) but lower recall (more missed true
transitions). The decline in recall is particularly pronounced at the fine level, where boundaries are
shorter and more numerous, making them more sensitive to neighborhood enlargement.

\end{document}